\documentclass[journal,10pt]{IEEEtran}
\usepackage[colorlinks,
linkcolor=blue,
anchorcolor=blue,
citecolor=blue,]{hyperref}

\usepackage{cite}
\usepackage{graphicx}
\usepackage{subfigure}
\usepackage{slashbox}
\makeatletter

\newcommand{\Rmnum}[1]{\expandafter\@slowromancap\romannumeral #1@}
\makeatother
\ifCLASSINFOpdf
\else
\fi

\usepackage{amssymb}

\usepackage{mathrsfs}
\usepackage{amsmath,bm}

\usepackage{graphicx}
\usepackage{caption}
\usepackage{algpseudocode}
\usepackage{multirow}
\usepackage{amssymb}
\usepackage{makecell}
\usepackage{booktabs}
\usepackage[ruled]{algorithm2e}

\begin{document}
%
\title{Discriminative Anchor Learning for Efficient Multi-view Clustering}

\author{Yalan~Qin,
	Nan~Pu,
	Hanzhou~Wu,
	and Nicu~Sebe
	\thanks{This work was supported by 2024 Xizang Autonomous Region Central Guided Local Science and Technology Development Fund Project under Grant Number XZ202401YD0002, and National Natural Science Foundation of China under Grant Number U23B2023, and China Postdoctoral Science Foundation under Grant 2023M742208. \emph{(Corresponding author: Nan Pu.)}}
	\thanks{Y. Qin and H. Wu are with the School of Communication and Information Engineering, Shanghai University, Shanghai 200444, China.}
	\thanks{N. Pu and N. Sebe are with the Dept. of Information Engineering and Computer Science, University of Trento, Trento, 38100, Italy.}}

\markboth{Submitted to IEEE TRANSACTIONS ON MULTIMEDIA}%
{Shell \MakeLowercase{\textit{et al.}}: Bare Demo of IEEEtran.cls for Journals}

\maketitle

\begin{abstract}
Multi-view clustering aims to study the complementary information across views and discover the underlying structure. For solving the relatively high computational cost for the existing approaches, works based on anchor have been presented recently. Even with acceptable clustering performance, these methods tend to map the original representation from multiple views into a fixed shared graph based on the original dataset. However, most studies ignore the discriminative property of the learned anchors, which ruin the representation capability of the built model. Moreover, the complementary information among anchors across views is neglected to be ensured by simply learning the shared anchor graph without considering the quality of view-specific anchors. In this paper, we propose discriminative anchor learning for multi-view clustering (DALMC) for handling the above issues. We learn discriminative view-specific feature representations according to the original dataset and build anchors from different views based on these representations, which increase the quality of the shared anchor graph. The discriminative feature learning and consensus anchor graph construction are integrated into a unified framework to improve each other for realizing the refinement. The optimal anchors from multiple views and the consensus anchor graph are learned with the orthogonal constraints. We give an iterative algorithm to deal with the formulated problem. Extensive experiments on different datasets show the effectiveness and efficiency of our method compared with other methods.
\end{abstract}


\IEEEpeerreviewmaketitle

\section{INTRODUCTION}
Data can be characterized from multiple modalities, which is usually called multi-view data in computer vision and data analysis \cite{Yuans24,Yuans23,Yangq23,Yangq231}. Different from clustering the data with single view \cite{Yala222,Yala211,Yala232,Yala233,Nanp23}, multi-view clustering aims to exploit the underlying structure and discover the complementary information among different views for clustering. As an effective method for performing the clustering procedure \cite{Peng23,Peng22,Wuw21,Jia23,Yala24,Chenyo22,Yala221,Yala231,Yala241,Yala234,Yala242,Xingf23,Qian18,Qian21,Qian211} on the data from different sources, the existing approaches can be classified into several categories including multiple kernel clustering \cite{Tie23}, methods based on matrix factorization \cite{Ghu19}, multi-view subspace clustering \cite{Hong15} and approaches built on the graph \cite{Weix22}. Multiple kernel clustering \cite{Liux21} maximizes the kernel coefficients and partition matrix to find a shared cluster assignment matrix. Methods based on matrix factorization \cite{Jing13} attempt to seek for a consensus matrix to achieve common information among views. Multi-view subspace clustering adopts a self-expressive framework to achieve a reconstruction matrix. Methods based on graph \cite{Xuel22} conduct eigen-decomposition upon the obtained Laplacian matrix, which usually rely on the spectral clustering and partition the data with the guidance of this clustering algorithm.

Even with acceptable performance, these methods encounter the relatively high computation cost, restricting their ability for the data with large scales. To solve the efficiency issue, sampling-based methods and researches for non-negative matrix factorization (NMF) are two of the representive hot topics \cite{Zhao20,Jing13}. Kang et al. \cite{Zhao20} achieved the anchor points based on $ K $-means and linearly reconstructed the original data point by adopting the similarity graph matrix. Li et al. \cite{Yeq15} selected independent anchors for different views and employed the corresponding anchor map to replace the original map. Liu et al. \cite{Suy22} fused the original data reconstruction and anchor learning into a unified framework for learning an optimal consensus graph. Methods based on NMF factorize the original feature representations into two different parts, i.e., view-specific base matrices and consensus coefficient matrix. Liu et al. \cite{Jiy21} gave a method based on matrix factorization and removed the restriction with the guidance of the non-negative constraints. Wang et al. \cite{Jing18} adopted a diversity term to decrease the redundancy and solved the formulated optimization problem.

As another kind of representive efficient methods for multi-view data, works built on the anchor graph have presented promising capability. These methods typically represent the data structure by generating smaller set of anchors from the original dataset and build an $ n\times l $ anchor graph instead of the $ n\times n $ graph, where $ l $ is the number of total anchors and $ n $ indicates the size of dataset. For instance, Wang et al. \cite{Siw22} employed multiple projection matrices and the set of latent consensus anchors for learning a unified anchor graph for different views. Li et al. \cite{Xue22} presented a novel scalable method based on the anchor graph fusion. Wang et al. \cite{Wangj23} introduced a novel approach based on the efficient multiple $ K $-means, which incorporates two approximated partition matrices rather than the original ones for each base kernel. Chen et al. \cite{Chenyo23} developed a scalable multi-view clustering framework by combining tensor learning and dynamic anchor learning (TDASC), which adopts the anchor learning to achieve smaller view-specific graphs. Efficient and effective one-step multiview clustering (E$ ^{2} $OMVC) is developed to directly achieve the clustering indicators without large-time burden and construct smaller similarity graphs from different views based on the anchor graphs \cite{Wangju23}. Huang et al. \cite{Huangbo20} adopt three attributes to construct the multi-kernel models for features with multiple channels. Some existing works based on the graph \cite{Meng21} either learn anchor graphs for different views and heuristically concatenate these anchor graphs or achieve a single anchor graph shared by all views. They usually regard spectral partitioning and anchor graph learning as two separated steps.
\begin{figure*}[!t]
	\begin{center}
		\includegraphics[width=0.88\textwidth]{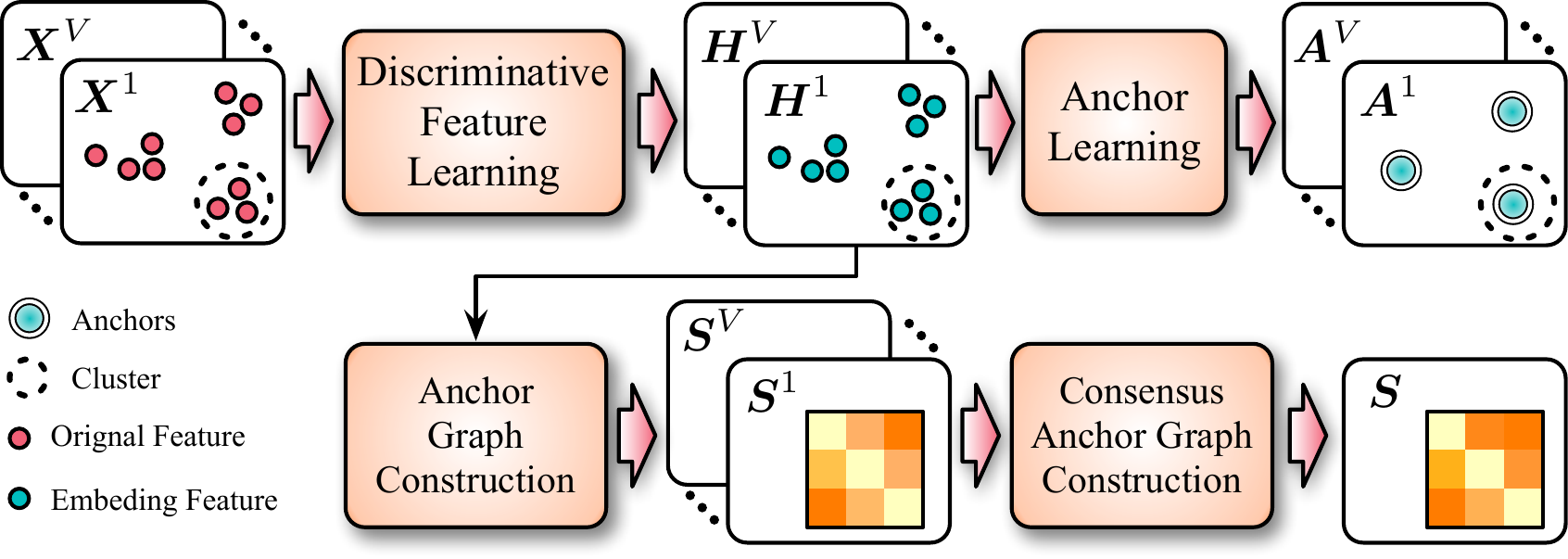}
		\caption{(a) Overview of our framework. We first learn discriminative feature representations of the original multi-view dataset, then the consensus anchor graph is built on these feature representations. Lastly, with the obtained consensus anchor graph, $ K $-means is employed to achieve the final clustering results.
			\label{fig:framework}}
	\end{center}
	\vspace{-2em}
\end{figure*}
Although the improved clustering performance can be achieved by these methods, we find that they usually map the original feature representations into a fixed shared anchor graph. However, data representations of different views often belong to specific anchors, and the final performance largely depends on the quality of view-specific anchors. Most studies ignore this issue and just learn the view-specific anchors based on the original feature representations from different views, which fail to ensure the discriminative property of the learned anchors and ruin the representation capability of the built model. Besides, the complementary knowledge among view-specific anchors from different views is not able to be guaranteed by directly learning the shared anchor graph without considering the quality of view-specific anchors. Then, the final shared anchor graph fails in extracting the complementary feature representations from multiple views. Moreover, the separated process between discriminative view-specific feature learning and consensus anchor graph construction in multi-view clustering may lead to the sub-optimal clustering results. Then these two parts will not be able to improve each other for reaching a refinement.

In this paper, we propose discriminative anchor learning for multi-view clustering (DALMC) to deal with the above issues, which is shown in Fig. 1. To increase the quality of the shared anchor graph and gain complementarity among multiple views, we learn discriminative feature representation of each view from the original multi-view dataset and then construct the consensus anchor graph with the guidance of these discriminative feature representations. The discriminative view-specific feature learning and consensus anchor graph construction are formulated into a unified framework, where these two parts are able to improve each other for realizing the refinement. The optimal anchors of each view and the consensus anchor graph are learned with the orthogonal constraints. Considering the existed differences among the qualities of view-specific feature representations, we automatically determine the importance of each one according to its contribution to the final clustering results. The major difference between TDASC and our method is that we consider the discriminative view-specific feature representation learning in obtaining consensus anchor graph. We show that DALMC has the linear time complexity, which enables it to efficiently deal with the large-scale dataset.

The major contributions in this paper are:

\begin{enumerate}
	\item We propose the discriminative anchor learning for multi-view clustering in this work, which is able to increase the discriminability of view-specific feature representations and achieve complementarity among different views. We learn discriminative view-specific feature representations and then construct a consensus anchor graph with these representations, which can remove the existing redundant information in feature representations from different views. The view-specific anchors are defined as orthogonal vectors and can be directly learned by optimization.
	
	\item We formulate the discriminative view-specific feature learning and consensus anchor graph construction into a unified framework, where these two parts improve each other to reach a refinement. The discriminative feature representations of the original dataset are achieved by imposing the orthogonal constraints on the base matrix in factorization. We learn the optimal anchors of each view and the consensus anchor graph based on the obtained discriminative feature representations.
	
	\item We give the alternate optimization algorithm for solving the formulated problem of DALMC. The linear time complexity enables DALMC for dealing with the datasets with large scales. Experiments on several datasets are performed to validate the efficiency and effectiveness of DAMLC compared with other methods.
\end{enumerate}

\section{Related Work}
In this part, we give an overview of subspace clustering for single-view and multi-view data, and then present the related works based on anchor graph.

\subsection{Subspace Clustering}
Given single-view dataset $ X\in R^{d\times n} $, subspace clustering expresses each data point by linearly combining the others and obtains the combination coefficient by minimizing the reconstruction loss, where $ d $ and $ n $ correspond to the dimension and size of dataset, respectively. The above process is formulated as:
\begin{equation}
\begin{split}
\begin{aligned}
\min_{S}\lVert X-XS\rVert_{F}^{2}+\lambda f(S),\;\;s.t.\;S\geq 0,\; S\bm{1}=1,
\end{aligned}
\end{split} 
\end{equation} 
where $ \lambda>0 $ represents the balance parameter, $ f(.) $ is the regularizer function, $ \bm{1} $ is a vector with each entry being one, $ S\in R^{n\times n} $ indicates the similarity matrix and each term in $ S $ is non-negative. The first term denotes the error in the reconstruction procedure and the second one is adopted for regularization. We can observe that $ S $ owns the size of $ n\times n $, which is a challenge of scalability to the adopted datasets with large scales.

It takes $ \mathcal{O}(n^{3}) $ computation complexity in obtaining graph $ S $, which is a burden on both storage and computation for the dataset with large scales. Moreover, the subsequent clustering step also needs extra computation cost, i.e., spectral clustering consumes $ \mathcal{O}(n^{3}) $ complexity. For multi-view dataset $ \{X^{p}\}_{p=1}^{v}  $ with $ d^{p} $ and
$X^{p}\in R^{d^{p}\times n}  $ being dimension and the data for the $ p $-th view, multi-view subspace clustering solves the problem as follows: 
\begin{equation}
\begin{split}
\begin{aligned}
\min_{S^{p}}\sum_{p=1}^{v}\lVert X^{p}-X^{p}S^{p}\rVert_{F}^{2}+\lambda f(S^{p}),\;\;s.t.\;S^{p}\geq 0,\; S^{p}\bm{1}=1.
\end{aligned}
\end{split} 
\end{equation} 
Most existing multi-view clustering approaches usually need at least $ \mathcal{O}(n^{2}k) $ complexity with $ k $ being the number of clusters, hence they are tend to be computational prohibitive. We then overview the methods of matrix factorization for single view and multiple views, respectively.

Give single-view dataset $ X $, clustering methods based on NMF factorize a data point into two non-negative parts, i.e., coefficient $ H\in R^{k\times n}$ and base $  Z\in R^{d\times k} $. NMF attempts to approximate $ X $ based on the product of $ Z $ and $ H $, which is formulated as
\begin{equation}
\begin{split}
\begin{aligned}
\min_{Z\geq 0,H\geq 0}f(X,ZH),
\end{aligned}
\end{split} 
\end{equation} 
where $ f(.) $ represents the loss function. The Frobenius norm is usually employed for measuring the loss between different items. By introducing the orthogonal constraints on the base matrix, we can rewrite Eq. (3) as follows:
\begin{equation}
\begin{split}
\begin{aligned}
\min_{Z\geq 0,H\geq 0}\lVert X-ZH\rVert_{F}^{2},\; s.t.\;Z^{T}Z=I.
\end{aligned}
\end{split} 
\end{equation} 
Considering that the non-negative constraint prevents from obtaining a more discriminative embedding $ H $, the clustering method based on matrix factorization can be achieved by removing this restriction \cite{Liux21}. It can be explained by the fact that removing the non-negativity constraint encourages the model to learn a more discriminative embedding in a larger search region \cite{Jiy21}. The above process is formulated as follows
\begin{equation}
\begin{split}
\begin{aligned}
\min_{Z,H}\lVert X-ZH\rVert_{F}^{2},\; s.t.\;Z^{T}Z=I.
\end{aligned}
\end{split} 
\end{equation} 
Based on the obtained $ H $, we then perform the traditional clustering method to gain the final cluster assignments, i.e., $ K $-means algorithm. After extending Eq. (5) for multi-view dataset, we can obtain 
\begin{equation}
\begin{split}
\begin{aligned}
\min_{Z^{p},H^{p}}\sum_{p=1}^{v}\lVert X^{p}-Z^{p}H^{p}\rVert_{F}^{2}+\lambda \psi(H^{p},Z^{p}),\; s.t.\;(Z^{p})^{T}Z^{p}=I,
\end{aligned}
\end{split} 
\end{equation} 
where $ \psi(.) $ is the regularization for $ Z$ and $ H $ of the $ p $-th view.

\subsection{Anchor Graph}
To reduce the computation complexity, some methods based on anchor have been given \cite{Yeq15} for datasets with large scales. They are able to focus on the $ n\times l $ anchor graph, where $ l\ll n $ is the total number of anchors. Nie et al. \cite{Fei23} learned a structured optimal anchor graph based on a dictionary matrix with the Laplacian rank constraint. Sun et al. \cite{Meng21} adopted the orthogonal and unified anchors for learning a unified anchor graph, where the weights of each view can be adaptively learned. A more flexible and representative cluster indicator and anchor representation can be obtained in the work \cite{Man22}. Zhang et al. \cite{Tiej22} reconstructed the base partition for each view based on anchor learning with the consensus reconstruction matrix and view-specific anchor. Liu et al. \cite{Suy22} imposed a graph connectivity constraint in anchor learning. Nie et al. \cite{Feip17} considered the similarity matrix for each view to learn a graph constrained by the Laplacian rank.

However, data representation of each view usually belongs to specific anchors, and the quality of view-specific anchors tends to influence the final clustering performance. Most of the existing works neglect this issue and just learn view-specific anchors based on the original representations from different views, which neglect the discriminative property of the learned anchors and influence the representation capability of the built model. Moreover, the complementary information among anchors across multiple views is not ensured by directly learning the shared anchor graph without considering the quality of view-specific anchors.

\section{The proposed method}
In this part, we first present the motivation and formulation of the proposed DALMC. Then the detailed optimization process and the complexity analysis of DALMC are also introduced in the following.
\subsection{Motivation and Formulation}
The existing studies have shown that adopting anchors to reduce the computation complexity is an effective way \cite{Yeq15}. Despite these methods have shown advantages, there are some shortcomings remained to be solved. First, the current methods merely learn view-specific anchors based on the original multi-view dataset, which ignore the discriminative property of the learned anchors and limit the representation capability of the model. Second, these methods pay few attention to guaranting the complementary information among anchors across multiple views with the guidance of high-quality view-specific anchors. To achieve the desired anchors for clustering, we learn discriminative view-specific feature representations and then construct a consensus anchor graph based on these representations. The discriminative view-specific feature learning and consensus anchor graph construction are formulated into a unified framework, where these two parts can improve each other to reach a refinement. To be specific, we obtain discriminative view-specific feature representation $ H^{p}\in R^{d'^{p}\times n} $ and basis matrix $ Z^{p}\in R^{d^{p}\times d'^{p}} $, where $ d'^{p} $ is the dimension of $ H^{p} $ and $ d'^{p}<d^{p} $. Then a consensus sampling matrix $ C\in R^{n\times l} $ is adopted to obtain $ l $ anchors by linearly combining $ n $ data points in $ H^{p} $, which reduces the number of data points. To guarantee that the consensus anchor graph and view-specific anchor graphs to be consistent, we enforce the consensus anchor graph to be close to the anchor graph from each view. The above process is formulated as
\begin{equation}
\begin{split}
\begin{aligned}
&\min_{Z^{p},H^{p},\alpha,C,S^{p},S}\sum_{p=1}^{v}\frac{1}{2}\alpha_{p}^{2}\lVert X^{p}-Z^{p}H^{p} \rVert_{F}^{2}+\gamma \sum_{p=1}^{v}\lVert S-S^{p}\rVert_{F}^{2}\\
&\;\;\;-\sum_{p=1}^{v}\beta Tr(H^{p}(H^{p}CS^{p})^{T}),\\
& s.t.\;Z^{p^{T}}Z^{p}=I,\;\alpha^{T}1=1,\;\alpha\geq 0,\;C^{T}C=I,\\
&\;\;\;\;\;\; 0\leq S\leq 1,\;0\leq S_{p}\leq 1,
\end{aligned}
\end{split} 
\end{equation} 
where $ \alpha_{p} $, $ \beta $ and $ \gamma $ are weight parameters for balancing different terms. $ S^{p} $ and $ S $ indicate view-specific and consensus anchor graphs, respectively. Note that the last term of the above formulation is obtained based on self-expressive formulation $ H^{p}=H^{p}CS^{p} $. The loss function of self-expressive formulation is based on the similarity between $ H^{p} $ and $H^{p}CS^{p}  $, which is denoted by $ Tr(H^{p}(H^{p}CS^{p})^{T}) $.
In order to improve the discriminability of view-specific feature representations, we introduce the orthogonal constraint on the basis matrix $ Z^{p} $ in Eq. (7). The quadratic term regarding $ H^{p} $ inevitably increases the computation complexity with the quadratic of the dataset size. Moreover, the quality of anchors is limited by $ H^{p} $'s representation capability by sampling $ H^{p} $ with $ C $, which limits the improvement of performance. We also find that $ S $ is learned by the fusion of $ S^{p} $ from different views.

To overcome the above shortcomings, we directly learn the anchor $ A^{p}\in R^{ d'^{p}\times l} $ for each view instead of adopting $ H^{p}C $, which is able to avoid using $ C $ for extracting the data points from $ H^{p} $. Then the quadratic term regarding $ H^{p} $ is removed in the optimization process, which reduces the computation complexity to linear. Besides, the anchor with high quality is not limited to $  H^{p} $ and we give each view with a different weight. Then, we can rewrite the formula as:
\begin{equation}
\begin{split}
\begin{aligned}
&\min_{Z^{p},H^{p},A^{p},\alpha,S^{p},S}\sum_{p=1}^{v}\frac{1}{2}\alpha_{p}^{2}\lVert X^{p}-Z^{p}H^{p} \rVert_{F}^{2}+ \sum_{p=1}^{v}\gamma\lVert S-S^{p}\rVert_{F}^{2}\\
&\;\;\;-\sum_{p=1}^{v}\beta Tr(H^{p}(A^{p}S^{p})^{T}),\\
& s.t.\;Z^{p^{T}}Z^{p}=I,\;\alpha^{T}1=1,\;\alpha\geq 0,\\
&\;\;\;\;\;\; 0\leq S\leq 1,\;0\leq S^{p}\leq 1,\;A^{p^{T}}A^{p}=I.
\end{aligned}
\end{split} 
\end{equation} 
In order to simplify the optimization process in Eq. (8), we impose the orthogonal constraint on $ H^{p} $ instead of $ Z^{p} $. It is observed that the dimension of $H_{p}  $ is usually smaller than that of $ Z_{p} $. Since $ S $ has fused multi-view information in reconstructing $ H^{p} $ for each view, it is not necessary to learn $ S^{p} $ and then adopt them to learn $ S $. To enhance the representation capability of the consensus matrix, we employ the orthogonal constraints for replacing the previous constraints on $ S $. Different from the orthogonality constraints imposed on the anchor achieved by the original dataset in \cite{Man22}, we learn discriminative feature representations and build anchors of each view with orthogonality constraints based on the obtained feature representations. We formulate the above process as
\begin{equation}
\begin{split}
\begin{aligned}
&\min_{Z^{p},H^{p},A^{p},\alpha,S}\sum_{p=1}^{v}\frac{1}{2}\alpha_{p}^{2}\lVert X^{p}-Z^{p}H^{p} \rVert_{F}^{2}
-\sum_{p=1}^{v}\beta Tr(H^{p}(A^{p}S)^{T}),\\
& s.t.\;H^{p^{T}}H^{p}=I,\;\alpha^{T}1=1,\;\alpha\geq 0,\\
&\; \;\;\;\;\;S^{T}S=I,\;A^{p^{T}}A^{p}=I.
\end{aligned}
\end{split} 
\end{equation} 
After obtaining the optimal $ S $, we use it as the input for $ K $-means to achieve the final clustering result. Compared with the work in \cite{Suy22}, the proposed method considers learning the discriminative feature representation of each view and constructs anchors based on these representations. Different from the work in \cite{Siw22}, our method learns view-specific feature representations with high quality and the consensus anchors in a unified framework.

\subsection{Optimization}
The problem in Eq. (9) is jointly nonconvex when all variables are simultaneously considered. Therefore, we adopt the alternative algorithm for optimizing each variable with the others being fixed.

$ H^{p} $-subproblem: With the other variables except for $ H^{p} $ being fixed, we reformulate the problem in Eq. (9) as:
\begin{equation}
\begin{split}
\begin{aligned}
&\min_{H^{p}}\sum_{p=1}^{v}\frac{1}{2}\alpha_{p}^{2}\lVert X^{p}-Z^{p}H^{p} \rVert_{F}^{2}
-\sum_{p=1}^{v}\beta Tr(H^{p}(A^{p}S)^{T}),\\
& s.t.\;H^{p^{T}}H^{p}=I.
\end{aligned}
\end{split} 
\end{equation} 
The above formulation can be transformed into
\begin{equation}
\begin{split}
\begin{aligned}
&\max_{H^{p}} Tr(H^{p}B),\\
&s.t.\;H^{p^{T}}H^{p}=I,
\end{aligned}
\end{split} 
\end{equation}
where $ B=\alpha_{p}^{2}\sum_{p=1}^{v}X^{p^{T}}Z^{p}+\beta S^{T}A^{p^{T}} $. Eq. (11) can be solved via SVD and the corresponding computation complexity is $ \mathcal{O}(nd^{p^{2}}) $.

$ Z^{p} $-subproblem: With the other variables except for $ Z^{p} $ being fixed, we reformulate the problem in Eq. (9) as:
\begin{equation}
\begin{split}
\begin{aligned}
\min_{Z^{p}}\lVert X^{p}-Z^{p}H^{p} \rVert_{F}^{2}.
\end{aligned}
\end{split} 
\end{equation}
We differentiate the above objective function regarding $ Z^{p} $ and set the derivative to be zero. Then $ Z^{p} $ is updated by
\begin{equation}
\begin{split}
\begin{aligned}
Z^{p}=X^{p}H^{p^{T}}.
\end{aligned}
\end{split} 
\end{equation}

$ S $-subproblem: With the other variables except for $ S $ being fixed, we reformulate the problem in Eq. (9) as:

\begin{equation}
\begin{split}
\begin{aligned}
&\max_{S} Tr(H^{p}(A^{p}S)^{T}),\\
& s.t.\;S^{T}S=I.
\end{aligned}
\end{split} 
\end{equation} 
Likewise, it can be solved by SVD and the computation cost of Eq. (14) is $ \mathcal{O}(nl^{2}) $.

$ A^{p} $-subproblem: With the other variables except for $ A^{p} $ being fixed, we reformulate the problem in Eq. (9) as:
\begin{equation}
\begin{split}
\begin{aligned}
&\max_{A^{p}} Tr(H^{p}(A^{p}S)^{T}),\\
& s.t.\;A^{p^{T}}A^{p}=I.
\end{aligned}
\end{split} 
\end{equation} 
As Eq. (14), the above problem is solved by SVD and the corresponding computation complexity is $ \mathcal{O}(d^{p}l^{2}) $.
\begin{table}[h] \scriptsize
	\caption{Details of datasets in the experiment. }
	\centering
	\tabcolsep=0.5cm
	\begin{tabular}{|c|c|c|c|}
		\toprule[0.5pt]
		\textbf{DataSet}&\textbf{Data points}&\textbf{Clusters}&\textbf{Views }\\
		\toprule[0.5pt]
		\textbf{AWA}&30475&50&6\\
		\textbf{Caltech-256}&30607&257&4\\
		\textbf{Flower17}&1360&17&7\\
		\textbf{MNIST}&60000&10&3\\
		\textbf{TinyImageNet}&100000&200&4\\
		\textbf{VGGFace2}&72283&200&4\\
		\textbf{YoutubeFace-50}&126054&50&4\\
		\toprule[0.5pt]
	\end{tabular}
	\label{}
\end{table}
\begin{table*}[h] \scriptsize
	\caption{Clustering results based on ACC (\%) on different datasets. ``-
		" indicates out of memory.}
	\centering
	\tabcolsep=0.28cm
	\begin{tabular}{|c|ccccccccc|}
		\toprule[0.5pt]
		\textbf{Dataset}&\textbf{AMGL}&\textbf{BMVC}&\textbf{SMVSC}&\textbf{OPMC}&\textbf{UOMVSC}&\textbf{FPMVSCAG}&\textbf{OMSC}&\textbf{AWMVC}&\textbf{Ours}\\
		\toprule[0.5pt]
		\textbf{AWA}&-&8.60$ \pm $0.05&9.15$ \pm $0.10&9.20$ \pm $0.02&-&9.00$ \pm $0.10&9.25$ \pm $0.00&9.40$ \pm $0.05&\textbf{9.80}$ \pm $\textbf{0.10}\\
		\textbf{Caltech-256}&-&8.58$ \pm $0.02&9.60$ \pm $0.05&11.00$ \pm $0.10&-&9.40$ \pm $0.06&12.00$ \pm $0.05&12.76$ \pm $0.01&\textbf{13.12}$ \pm $\textbf{0.00}\\
		\textbf{Flower17}&9.60$ \pm $0.08&26.90$ \pm $0.05&27.90$ \pm $0.02&32.10$ \pm $0.00&36.30$ \pm $0.05&26.00$ \pm $0.01&47.50$ \pm $0.00&49.10$ \pm $0.05&\textbf{50.10}$ \pm $\textbf{0.02}\\
		\textbf{MNIST}&-&45.90$ \pm $0.05&98.72$ \pm $0.01&98.56$ \pm $0.03&-&98.80$ \pm $0.05&98.85$ \pm $0.00&98.82$ \pm $0.05&\textbf{98.90}$ \pm $\textbf{0.01}\\
		\textbf{TinyImageNet}&-&4.00$ \pm $0.00&3.05$ \pm $0.02&5.10$ \pm $0.05&-&2.92$ \pm $0.00&5.05$ \pm $0.00&5.12$ \pm $0.02&\textbf{5.40}$ \pm $\textbf{0.03}\\
		\textbf{VGGFace2}&-&3.90$ \pm $0.05&3.02$ \pm $0.03&3.85$ \pm $0.03&-&3.13$ \pm $0.02&6.72$ \pm $0.00&6.80$ \pm $0.02&\textbf{6.82}$ \pm $\textbf{0.05}\\
		\textbf{YoutubeFace-50}&-&66.00$ \pm $0.02&67.70$ \pm $0.05&69.30$ \pm $0.05&-&66.30$ \pm $0.01&75.00$ \pm $0.00&75.50$ \pm $0.04&\textbf{76.00}$ \pm $\textbf{0.02}\\
		\toprule[0.5pt]
	\end{tabular}
	\label{}
\end{table*}

\begin{table*}[h] \scriptsize
	\caption{Clustering results based on NMI (\%) on different datasets. ``-
		" indicates out of memory.}
	\centering
	\tabcolsep=0.28cm
	\begin{tabular}{|c|ccccccccc|}
		\toprule[0.5pt]
		\textbf{Dataset}&\textbf{AMGL}&\textbf{BMVC}&\textbf{SMVSC}&\textbf{OPMC}&\textbf{UOMVSC}&\textbf{FPMVSCAG}&\textbf{OMSC}&\textbf{AWMVC}&\textbf{Ours}\\
		\toprule[0.5pt]
		\textbf{AWA}&-&11.92$ \pm $0.03&10.70$ \pm $0.02&12.20$ \pm $0.05&-&10.80$ \pm $0.05&11.00$ \pm $0.00&11.30$ \pm $0.02&\textbf{12.00}$ \pm $\textbf{0.05}\\
		\textbf{Caltech-256}&-&31.80$ \pm $0.05&24.45$ \pm $0.04&32.75$ \pm $0.05&-&22.00$ \pm $0.05&32.50$ \pm $0.00&34.00$ \pm $0.02&\textbf{35.20}$ \pm $\textbf{0.01}\\
		\textbf{Flower17}&10.26$ \pm $0.02&25.60$ \pm $0.01&25.15$ \pm $0.03&29.70$ \pm $0.05&34.91$ \pm $0.05&26.00$ \pm $0.02&45.50$ \pm $0.00&49.70$ \pm $0.05&\textbf{51.00}$ \pm $\textbf{0.01}\\
		\textbf{MNIST}&-&39.55$ \pm $0.05&96.20$ \pm $0.01&95.82$ \pm $0.05&-&96.50$ \pm $0.04&95.90$ \pm $0.00&96.42$ \pm $0.05&\textbf{97.20}$ \pm $\textbf{0.01}\\
		\textbf{TinyImageNet}&-&13.72$ \pm $0.04&10.00$ \pm $0.05&\textbf{16.10}$ \pm $\textbf{0.02}&-&10.20$ \pm $0.05&13.75$ \pm $0.00&14.60$ \pm $0.01&15.50$ \pm $0.10\\
		\textbf{VGGFace2}&-&15.00$ \pm $0.05&10.32$ \pm $0.02&13.30$ \pm $0.05&-&9.52$ \pm $0.06&16.50$ \pm $0.00&18.56$ \pm $0.02&\textbf{19.00}$ \pm $\textbf{0.05}\\
		\textbf{YoutubeFace-50}&-&82.20$ \pm $0.01&82.75$ \pm $0.00&82.37$ \pm $0.01&-&83.50$ \pm $0.00&84.80$ \pm $0.00&85.98$ \pm $0.01&\textbf{86.27}$ \pm $\textbf{0.00}\\
		\toprule[0.5pt]
	\end{tabular}
	\label{}
\end{table*}

\begin{table*}[h] \scriptsize
	\caption{Clustering results based on F1-score (\%) on different datasets. ``-
		" indicates out of memory.}
	\centering
	\tabcolsep=0.28cm
	\begin{tabular}{|c|ccccccccc|}
		\toprule[0.5pt]
		\textbf{Dataset}&\textbf{AMGL}&\textbf{BMVC}&\textbf{SMVSC}&\textbf{OPMC}&\textbf{UOMVSC}&\textbf{FPMVSCAG}&\textbf{OMSC}&\textbf{AWMVC}&\textbf{Ours}\\
		\toprule[0.5pt]
		\textbf{AWA}&-&4.30$ \pm $0.02&6.14$ \pm $0.01&2.45$ \pm $0.02&-&6.42$ \pm $0.00&4.00$ \pm $0.00&4.54$ \pm $0.02&\textbf{6.50}$ \pm $\textbf{0.01}\\
		\textbf{Caltech-256}&-&6.25$ \pm $0.01&5.09$ \pm $0.00&9.25$ \pm $0.02&-&5.62$ \pm $0.03&10.20$ \pm $0.00&11.02$ \pm $0.01&\textbf{11.50}$ \pm $\textbf{0.00}\\
		\textbf{Flower17}&11.50$ \pm $0.02&16.60$ \pm $0.00&16.00$ \pm $0.01&12.86$ \pm $0.02&19.65$ \pm $0.05&16.63$ \pm $0.02&32.85$ \pm $0.00&34.50$ \pm $0.01&\textbf{36.00}$ \pm $\textbf{0.00}\\
		\textbf{MNIST}&-&33.58$ \pm $0.02&97.50$ \pm $0.01&96.82$ \pm $0.03&-&97.70$ \pm $0.01&97.50$ \pm $0.00&97.72$ \pm $0.00&\textbf{97.90}$ \pm $\textbf{0.00}\\
		\textbf{TinyImageNet}&-&1.56$ \pm $0.01&1.53$ \pm $0.00&1.32$ \pm $0.02&-&1.74$ \pm $0.01&1.65$ \pm $0.00&1.70$ \pm $0.00&\textbf{1.80}$ \pm $\textbf{0.00}\\
		\textbf{VGGFace2}&-&1.45$ \pm $0.00&1.46$ \pm $0.01&0.75$ \pm $0.03&-&1.45$ \pm $0.01&2.20$ \pm $0.05&2.45$ \pm $0.02&\textbf{2.70}$ \pm $\textbf{0.01}\\
		\textbf{YoutubeFace-50}&-&57.00$ \pm $0.08&60.74$ \pm $0.00&62.09$ \pm $0.05&-&62.00$ \pm $0.01&68.50$ \pm $0.00&70.50$ \pm $0.00&\textbf{72.00}$ \pm $\textbf{0.00}\\
		\toprule[0.5pt]
	\end{tabular}
	\label{}
\end{table*}

\begin{table*}[h] \scriptsize
	\caption{Clustering results based on Purity (\%) on different datasets. ``-
		" indicates out of memory.}
	\centering
	\tabcolsep=0.28cm
	\begin{tabular}{|c|ccccccccc|}
		\toprule[0.5pt]
		\textbf{Dataset}&\textbf{AMGL}&\textbf{BMVC}&\textbf{SMVSC}&\textbf{OPMC}&\textbf{UOMVSC}&\textbf{FPMVSCAG}&\textbf{OMSC}&\textbf{AWMVC}&\textbf{Ours}\\
		\toprule[0.5pt]
		\textbf{AWA}&-&10.90$ \pm $0.05&10.00$ \pm $0.02&11.20$ \pm $0.01&-&9.70$ \pm $0.00&10.95$ \pm $0.00&11.50$ \pm $0.01&\textbf{12.20}$ \pm $\textbf{0.02}\\
		\textbf{Caltech-256}&-&14.92$ \pm $0.01&11.45$ \pm $0.02&16.95$ \pm $0.00&-&11.00$ \pm $0.05&18.00$ \pm $0.00&18.80$ \pm $0.03&\textbf{20.00}$ \pm $\textbf{0.02}\\
		\textbf{Flower17}&10.75$ \pm $0.01&29.90$ \pm $0.00&29.30$ \pm $0.05&33.58$ \pm $0.02&37.92$ \pm $0.01&27.42$ \pm $0.01&48.50$ \pm $0.00&51.18$ \pm $0.00&\textbf{53.00}$ \pm $\textbf{0.02}\\
		\textbf{MNIST}&-&47.65$ \pm $0.02&98.75$ \pm $0.01&98.60$ \pm $0.02&-&98.85$ \pm $0.01&98.65$ \pm $0.05&98.87$ \pm $0.02&\textbf{99.00}$ \pm $\textbf{0.01}\\
		\textbf{TinyImageNet}&-&4.70$ \pm $0.01&3.20$ \pm $0.02&5.85$ \pm $0.05&-&2.96$ \pm $0.01&5.25$ \pm $0.00&5.78$ \pm $0.01&\textbf{5.90}$ \pm $\textbf{0.02}\\
		\textbf{VGGFace2}&-&4.68$ \pm $0.01&3.17$ \pm $0.02&4.45$ \pm $0.01&-&3.20$ \pm $0.02&6.58$ \pm $0.00&7.59$ \pm $0.01&\textbf{7.80}$ \pm $\textbf{0.05}\\
		\textbf{YoutubeFace-50}&-&73.65$ \pm $0.00&71.05$ \pm $0.05&72.56$ \pm $0.03&-&69.30$ \pm $0.02&75.50$ \pm $0.00&79.30$ \pm $0.01&\textbf{81.00}$ \pm $\textbf{0.00}\\
		\toprule[0.5pt]
	\end{tabular}
	\label{}
\end{table*}

\begin{table*}[h] \scriptsize
	\caption{Ablation study of the proposed method on different datasets in terms of ACC. }
	\centering
	\tabcolsep=1.6cm
	\begin{tabular}{|c|c|c|c|}
		\toprule[0.5pt]
		\textbf{DataSet}&\textbf{ablation-1}&\textbf{ablation-2}&\textbf{Ours}\\
		\toprule[0.5pt]
		\textbf{AWA}&9.42&9.36&\textbf{9.80}\\
		\textbf{Caltech-256}&12.80&12.75&\textbf{13.12}\\
		\textbf{Flower17}&48.60&49.25&\textbf{50.10}\\
		\textbf{MNIST}&97.50&98.20&\textbf{98.90}\\
		\textbf{TinyImageNet}&5.05&5.30&\textbf{5.40}\\
		\textbf{VGGFace2}&6.53&6.38&\textbf{6.82}\\
		\textbf{YoutubeFace-50}&74.70&73.53&\textbf{76.00}\\
		\toprule[0.5pt]
	\end{tabular}
	\label{}
\end{table*}

\begin{table*}[h] \scriptsize
	\caption{Ablation study of the proposed method on different datasets in terms of NMI. }
	\centering
	\tabcolsep=1.6cm
	\begin{tabular}{|c|c|c|c|}
		\toprule[0.5pt]
		\textbf{DataSet}&\textbf{ablation-1}&\textbf{ablation-2}&\textbf{Ours}\\
		\toprule[0.5pt]
		\textbf{AWA}&11.39&10.75&\textbf{12.00}\\
		\textbf{Caltech-256}&34.60&34.45&\textbf{35.20}\\
		\textbf{Flower17}&50.16&50.75&\textbf{51.00}\\
		\textbf{MNIST}&96.50&97.05&\textbf{97.20}\\
		\textbf{TinyImageNet}&14.35&15.24&\textbf{15.50}\\
		\textbf{VGGFace2}&18.43&18.29&\textbf{19.00}\\
		\textbf{YoutubeFace-50}&85.78&85.84&\textbf{86.27}\\
		\toprule[0.5pt]
	\end{tabular}
	\label{}
\end{table*}

\begin{table*}[h] \scriptsize
	\caption{Ablation study of the proposed method on different datasets in terms of F1-score. }
	\centering
	\tabcolsep=1.6cm
	\begin{tabular}{|c|c|c|c|}
		\toprule[0.5pt]
		\textbf{DataSet}&\textbf{ablation-1}&\textbf{ablation-2}&\textbf{Ours}\\
		\toprule[0.5pt]
		\textbf{AWA}&6.24&6.05&\textbf{6.50}\\
		\textbf{Caltech-256}&10.78&11.32&\textbf{11.50}\\
		\textbf{Flower17}&34.69&35.20&\textbf{36.00}\\
		\textbf{MNIST}&96.87&97.50&\textbf{97.90}\\
		\textbf{TinyImageNet}&1.60&1.73&\textbf{1.80}\\
		\textbf{VGGFace2}&2.56&2.39&\textbf{2.70}\\
		\textbf{YoutubeFace-50}&70.80&71.62&\textbf{72.00}\\
		\toprule[0.5pt]
	\end{tabular}
	\label{}
\end{table*}

\begin{table*}[h] \scriptsize
	\caption{Ablation study of the proposed method on different datasets in terms of Purity. }
	\centering
	\tabcolsep=1.6cm
	\begin{tabular}{|c|c|c|c|}
		\toprule[0.5pt]
		\textbf{DataSet}&\textbf{ablation-1}&\textbf{ablation-2}&\textbf{Ours}\\
		\toprule[0.5pt]
		\textbf{AWA}&11.80&11.56&\textbf{12.20}\\
		\textbf{Caltech-256}&18.74&19.30&\textbf{20.00}\\
		\textbf{Flower17}&51.80&52.58&\textbf{53.00}\\
		\textbf{MNIST}&98.65&98.70&\textbf{99.00}\\
		\textbf{TinyImageNet}&5.74&5.81&\textbf{5.90}\\
		\textbf{VGGFace2}&7.67&7.48&\textbf{7.80}\\
		\textbf{YoutubeFace-50}&80.60&80.73&\textbf{81.00}\\
		\toprule[0.5pt]
	\end{tabular}
	\label{}
\end{table*}
$ \alpha $-subproblem: With the other variables except for $ \alpha $ being fixed, we reformulate the problem in Eq. (9) as:

\begin{equation}
\begin{split}
\begin{aligned}
&\min_{\alpha}\sum_{p=1}^{v}\alpha_{p}^{2}\lVert X^{p}-Z^{p}H^{p} \rVert_{F}^{2},\\
& s.t.\;\alpha^{T}1=1,\;\alpha\geq 0.
\end{aligned}
\end{split} 
\end{equation} 
According to Cauchy-Schwarz inequality, we update $ \alpha_{p} $ by

\begin{equation}
\begin{split}
\begin{aligned}
\alpha_{p}=\frac{\frac{1}{\lVert X^{p}-Z^{p}H^{p} \rVert_{F}^{2}}}{\sum_{p=1}^{v}\frac{1}{\lVert X^{p}-Z^{p}H^{p} \rVert_{F}^{2}}}.
\end{aligned}
\end{split} 
\end{equation} 

We summarize the optimization process of the proposed method in Algorithm 1. To extend the proposed method for incomplete multi-view data, we can use the indicator matrix $ B_{p}\in\{0,1\}^{n\times n_{p}} $ to mark the unavailable data in the final optimal function. Then $X^{p}B_{p}  $ represents the available data points for the $ p $-th view.

\begin{algorithm}[!ht]
	\caption{Algorithm of the proposed method}
	
	\KwIn{Dataset $\left\lbrace X^{p}\right\rbrace _{p=1}^{v} $, cluster number $ k $.}
	\KwOut{The consensus coefficient $ S $.}
	\SetKwInOut{Initialize}{Initialize}
	\Initialize {Initialize $ Z^{p},\;H^{p},\;A^{p},\;\alpha,\;S $.}
	\Repeat{convergence}{
		Update $ H^{p} $ by Eq. (11);\\
		Update $  Z^{p} $ via Eq. (13);\\
		Update $ S$ by Eq. (14);\\
		Update $ A^{p}$ via Eq. (15);\\
		Update $ \alpha $ by Eq. (17);\\
	}
\end{algorithm}

\subsection{Complexity Analysis}
The computation complexity to update $ H^{p} $, $ S $ and $ A^{p} $ is $ \mathcal{O}(nd^{p^{2}}) $, $ \mathcal{O}(nl^{2}) $ and $ \mathcal{O}(d^{p}l^{2}) $, respectively. To obtain the optimal $ Z^{p} $, it costs $ \mathcal{O}(d^{p}n) $ to perform matrix multiplication. When updating $ \alpha_{p} $, it just needs $ \mathcal{O}(1) $. Then, the computation complexity of our method is $ \mathcal{O}(nl^{2}+\sum_{p=1}^{v}(d^{p}l^{2}+nd^{p^{2}}+d^{p}n)) $ at each iteration, which is linear to the size of multi-view dataset $ n $.

\subsection{Convergence Analysis}
The objective value in Eq. (9) monotonically decreases when each variable updates with the others being fixed. Thus, we just need to show that Eq. (9) has a lower bound in the proving process. According to Cauchy-Schwarz inequality, we have
\begin{equation}
\begin{split}
\begin{aligned}
Tr(H^{p}(A^{p}S)^{T})\leq \lVert H^{p}\rVert_{F}\lVert S^{T}\rVert_{F}\lVert A^{p^{T}}\rVert_{F}.
\end{aligned}
\end{split} 
\end{equation} 
Based on $ \beta\leq 1 $ and the above analysis, we have
\begin{equation}
\begin{split}
\begin{aligned}
&\beta Tr(H^{p}(A^{p}S)^{T})\leq Tr(H^{p}(A^{p}S)^{T})\\
&\leq \lVert H^{p}\rVert_{F}\lVert S^{T}\rVert_{F}\lVert A^{p^{T}}\rVert_{F}=l\sqrt{d^{p}}.
\end{aligned}
\end{split} 
\end{equation} 
Therefore, we can obtain a lower bound for Eq. (9) as follows:
\begin{equation}
\begin{split}
\begin{aligned}
&\sum_{p=1}^{v}\frac{1}{2}\alpha_{p}^{2}\lVert X^{p}-Z^{p}H^{p} \rVert_{F}^{2}
-\sum_{p=1}^{v}\beta Tr(H^{p}(A^{p}S)^{T}),\\
& \geq 0-l\sqrt{d^{p}}=-l\sqrt{d^{p}}.
\end{aligned}
\end{split} 
\end{equation} 
Since the objective value monotonically decreases for each iteration in the alternating optimization procedure, we can conclude that the algorithm is theoretically converged. The convergence our method is analyzed in the experiment.
\begin{figure*} [!htbp]
	\centering    
	
	\subfigure[AWA] 
	{
		\begin{minipage}[t]{0.23\linewidth}
			\centering          
			\includegraphics[width=1.8in]{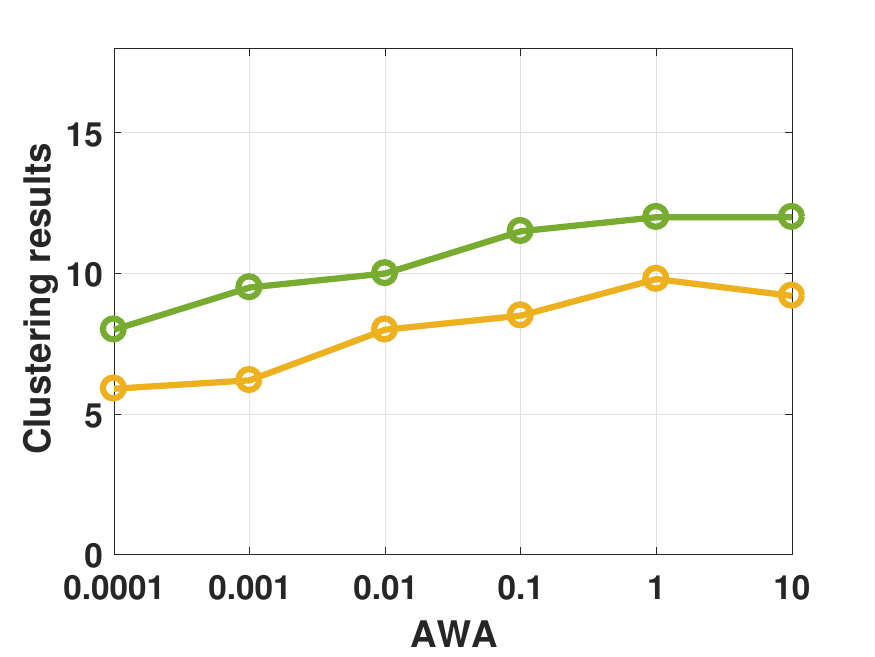}  
		\end{minipage}%
	}
	\subfigure[Caltech-256] 
	{
		\begin{minipage}[t]{0.23\linewidth}
			\centering      
			\includegraphics[width=1.8in]{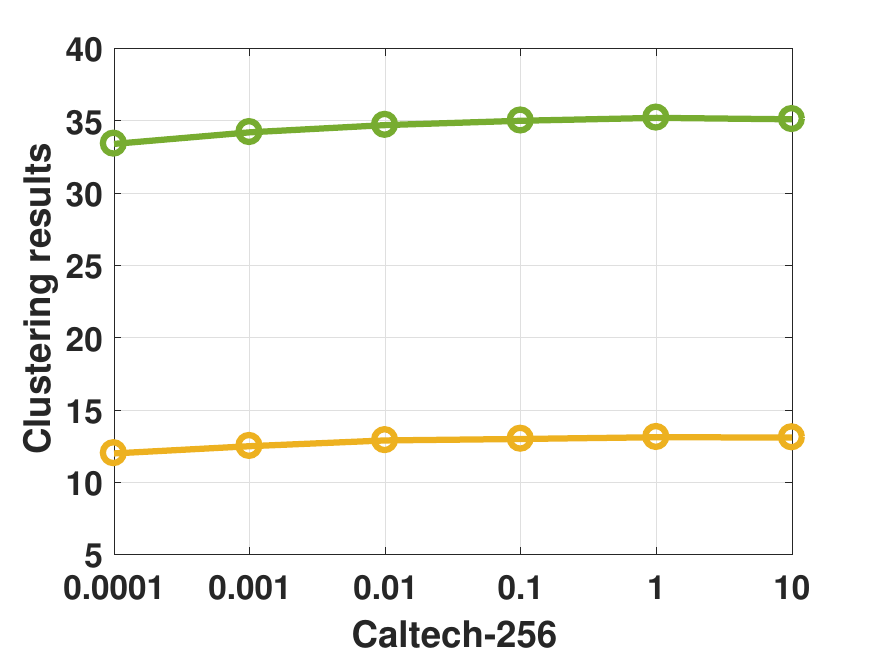} 
		\end{minipage}
	}	
	\subfigure[Flower17] 
	{
		\begin{minipage}[t]{0.23\linewidth}
			\centering      
			\includegraphics[width=1.8in]{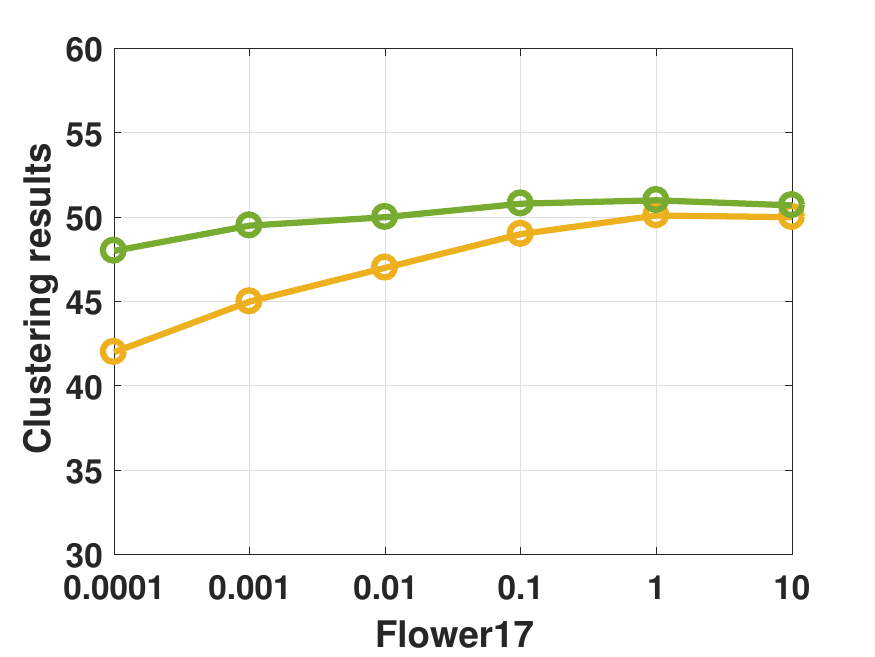} 
		\end{minipage}
	}	
	\subfigure[MNIST] 
	{
		\begin{minipage}[t]{0.23\linewidth}
			\centering          
			\includegraphics[width=1.8in]{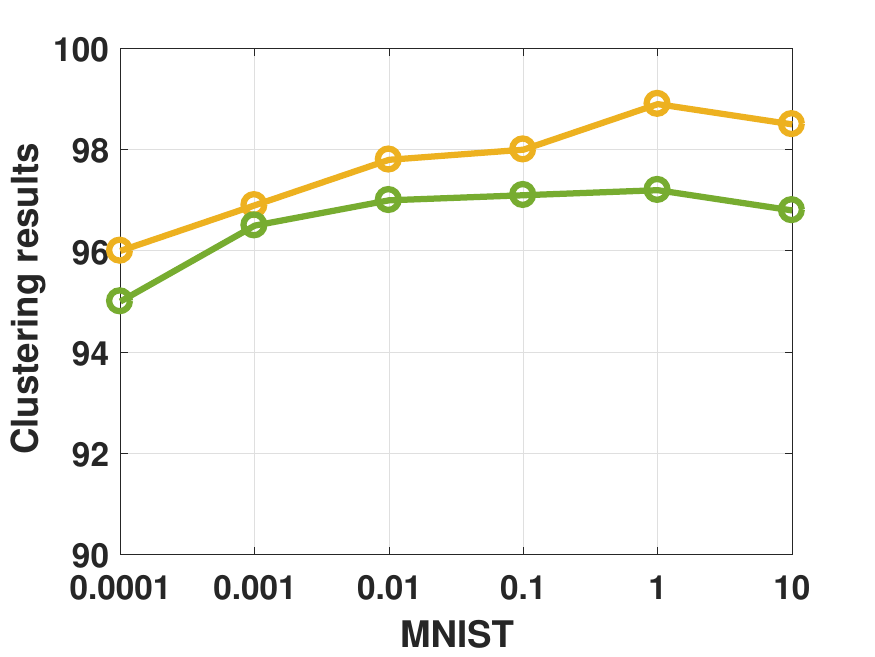}  
		\end{minipage}%
	}
	
	\subfigure[TinyImageNet] 
	{
		\begin{minipage}[t]{0.23\linewidth}
			\centering          
			\includegraphics[width=1.8in]{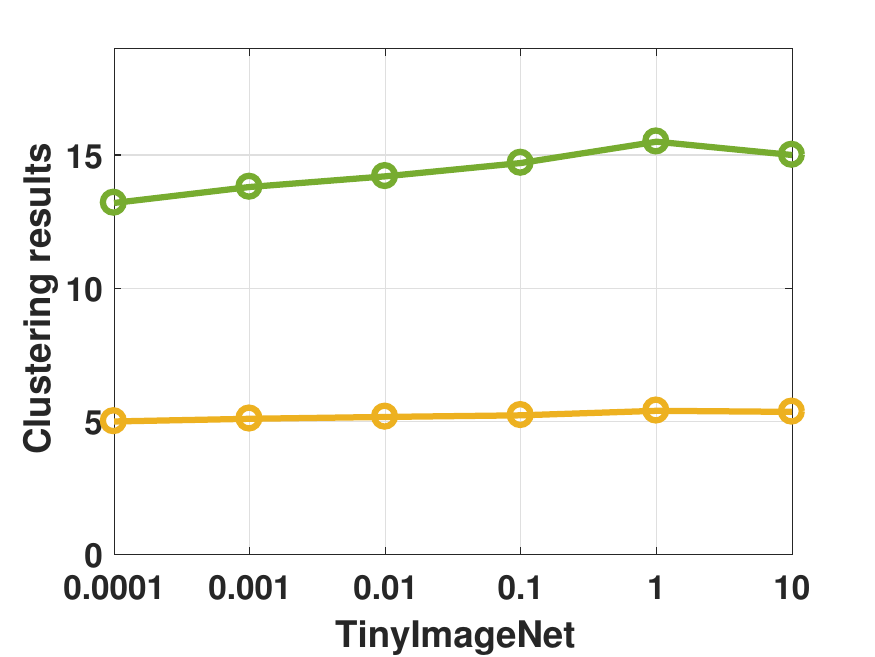}  
		\end{minipage}%
	}
	\subfigure[VGGFace2] 
	{
		\begin{minipage}[t]{0.23\linewidth}
			\centering      
			\includegraphics[width=1.8in]{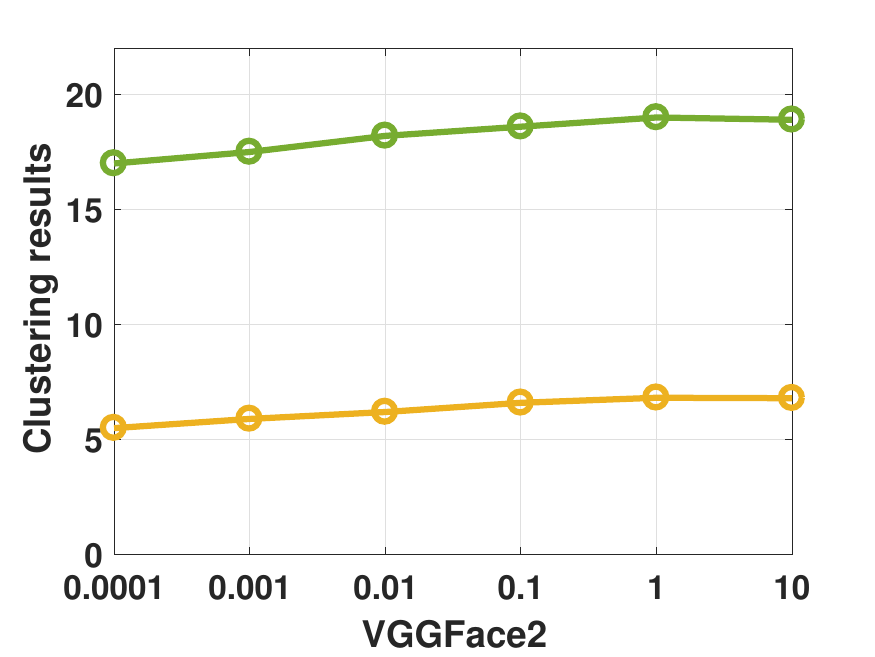} 
		\end{minipage}
	}	
	\subfigure[YoutubeFace-50] 
	{
		\begin{minipage}[t]{0.23\linewidth}
			\centering      
			\includegraphics[width=1.8in]{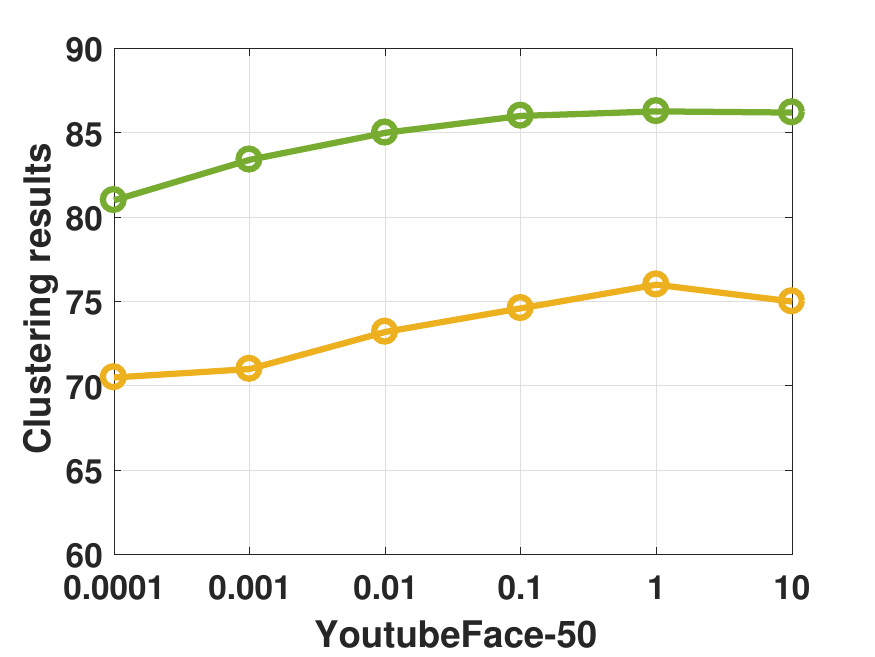} 
		\end{minipage}
	}
	\subfigure 
	{
		\begin{minipage}[t]{0.23\linewidth}
			\centering      
			\includegraphics[width=0.5in]{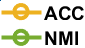} 
		\end{minipage}
	}	
	\caption{Parameter investigation of $ \beta $ on all datasets in terms of ACC and NMI.} 
	\label{fig1}  
\end{figure*}

\begin{figure*} [!htbp]
	\centering    
	
	\subfigure[AWA] 
	{
		\begin{minipage}[t]{0.23\linewidth}
			\centering          
			\includegraphics[width=1.8in]{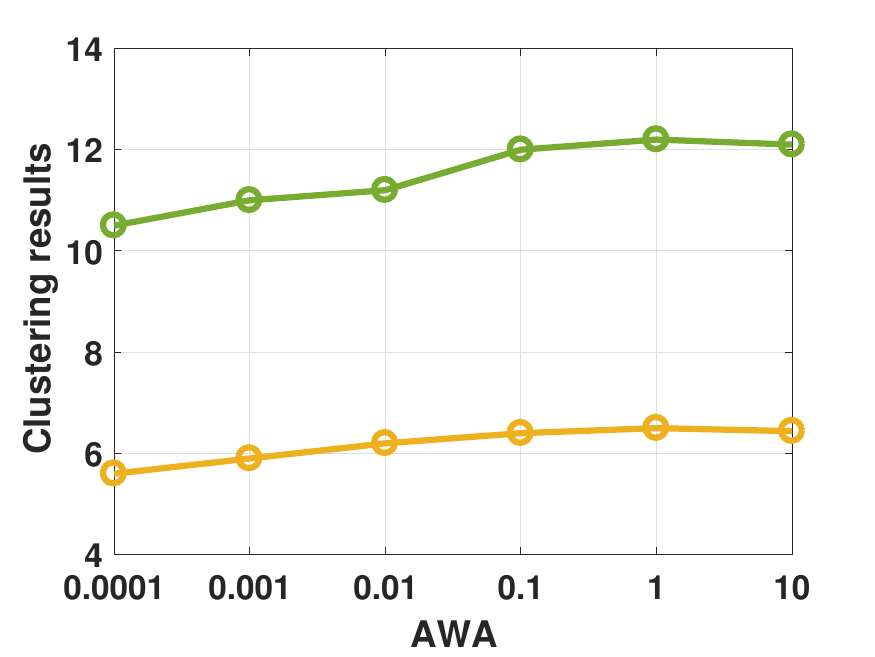}  
		\end{minipage}%
	}
	\subfigure[Caltech-256] 
	{
		\begin{minipage}[t]{0.23\linewidth}
			\centering      
			\includegraphics[width=1.8in]{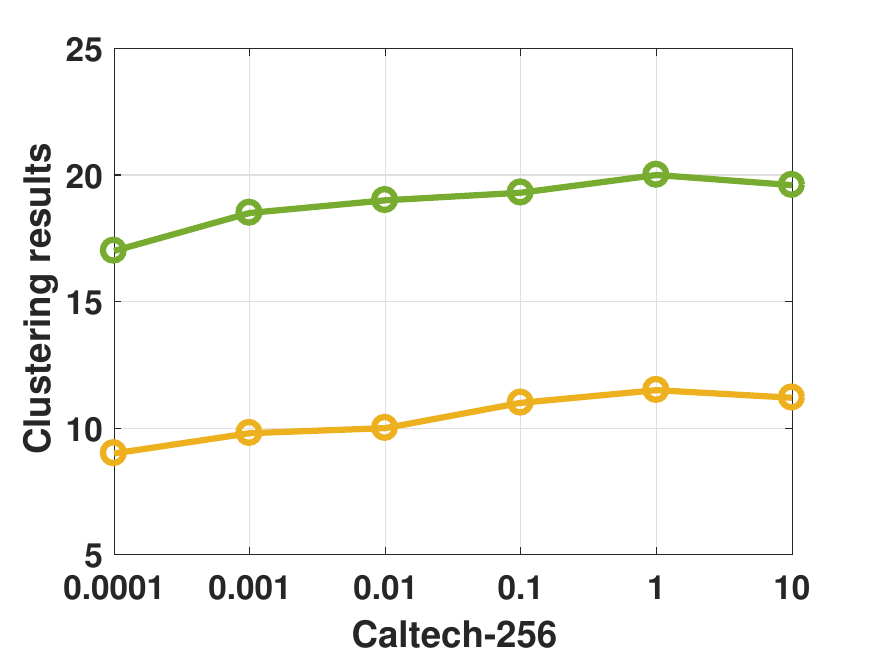} 
		\end{minipage}
	}	
	\subfigure[Flower17] 
	{
		\begin{minipage}[t]{0.23\linewidth}
			\centering      
			\includegraphics[width=1.8in]{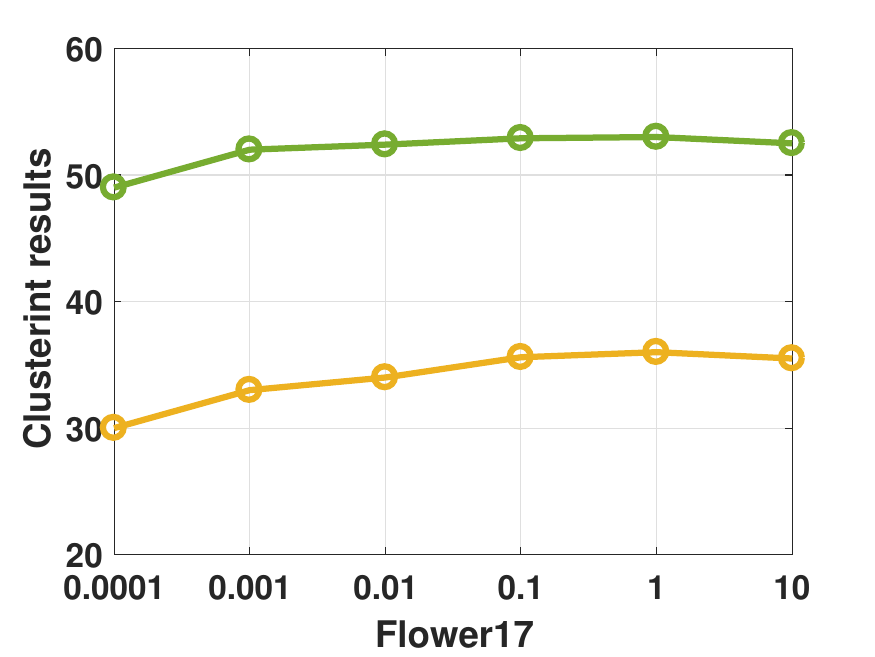} 
		\end{minipage}
	}	
	\subfigure[MNIST] 
	{
		\begin{minipage}[t]{0.23\linewidth}
			\centering          
			\includegraphics[width=1.8in]{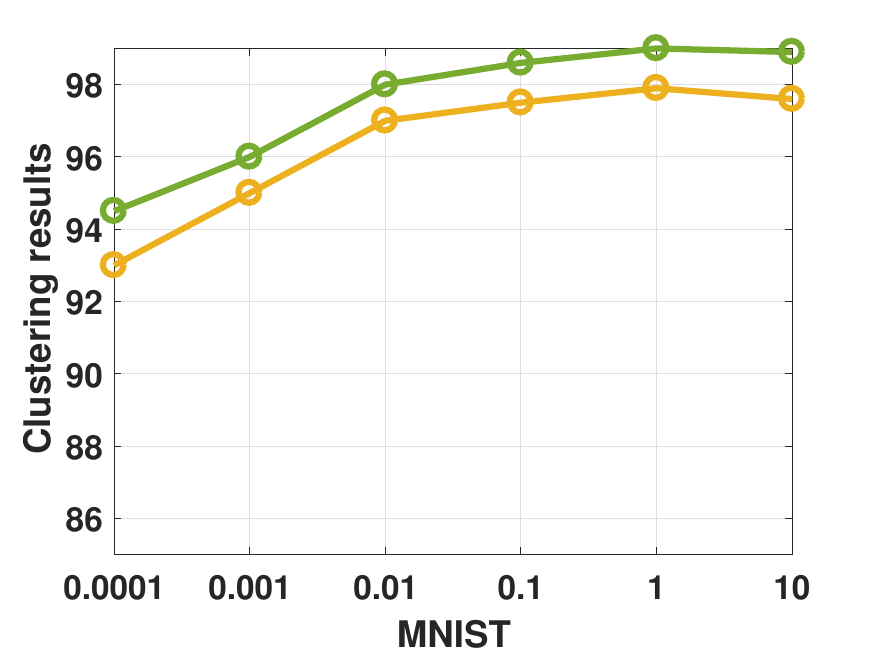}  
		\end{minipage}%
	}
	
	\subfigure[TinyImageNet] 
	{
		\begin{minipage}[t]{0.23\linewidth}
			\centering          
			\includegraphics[width=1.8in]{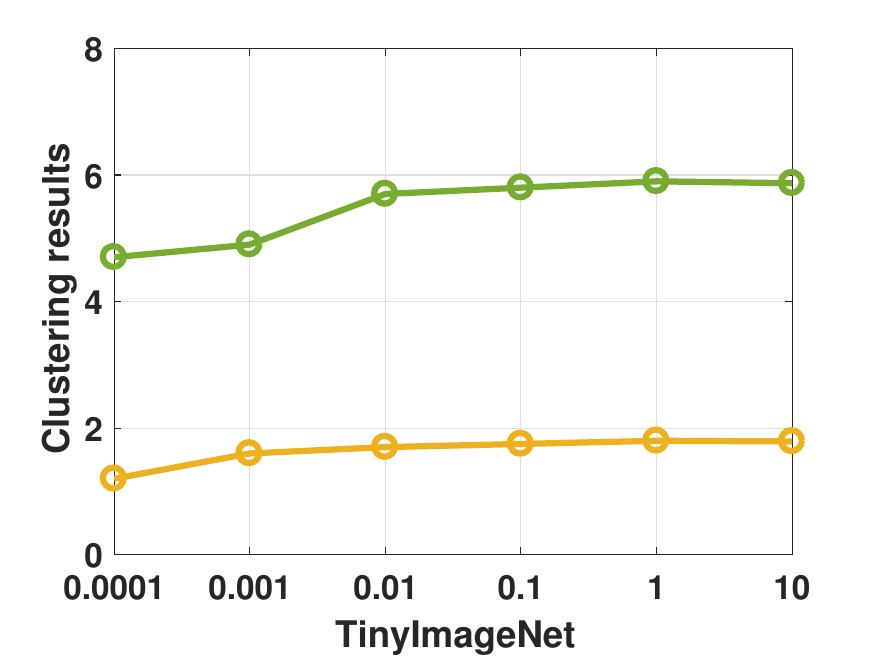}  
		\end{minipage}%
	}
	\subfigure[VGGFace2] 
	{
		\begin{minipage}[t]{0.23\linewidth}
			\centering      
			\includegraphics[width=1.8in]{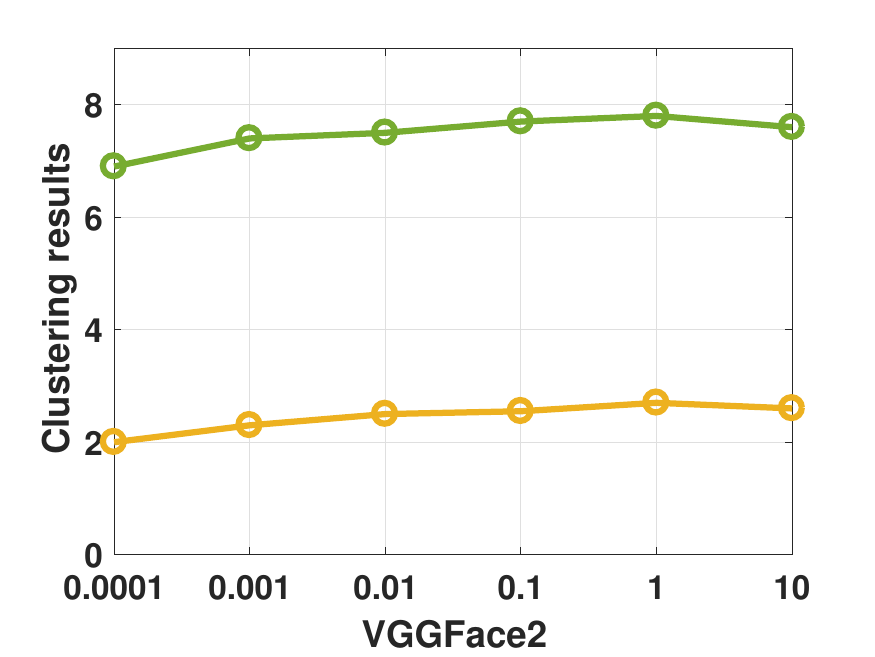} 
		\end{minipage}
	}	
	\subfigure[YoutubeFace-50] 
	{
		\begin{minipage}[t]{0.23\linewidth}
			\centering      
			\includegraphics[width=1.8in]{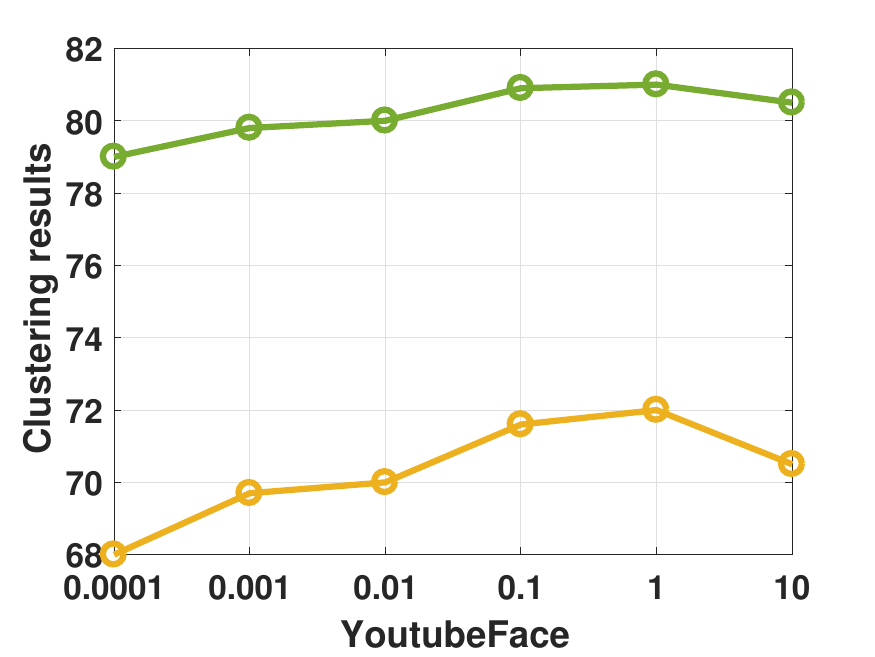} 
		\end{minipage}
	}
	\subfigure 
	{
		\begin{minipage}[t]{0.23\linewidth}
			\centering      
			\includegraphics[width=0.7in]{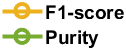} 
		\end{minipage}
	}	
	\caption{Parameter investigation of $ \beta $ on all datasets in terms of F1-score and Purity.} 
	\label{fig1}  
\end{figure*}

\begin{figure*} [!htbp]
	\centering    
	
	\subfigure[AWA] 
	{
		\begin{minipage}[t]{0.23\linewidth}
			\centering          
			\includegraphics[width=1.8in]{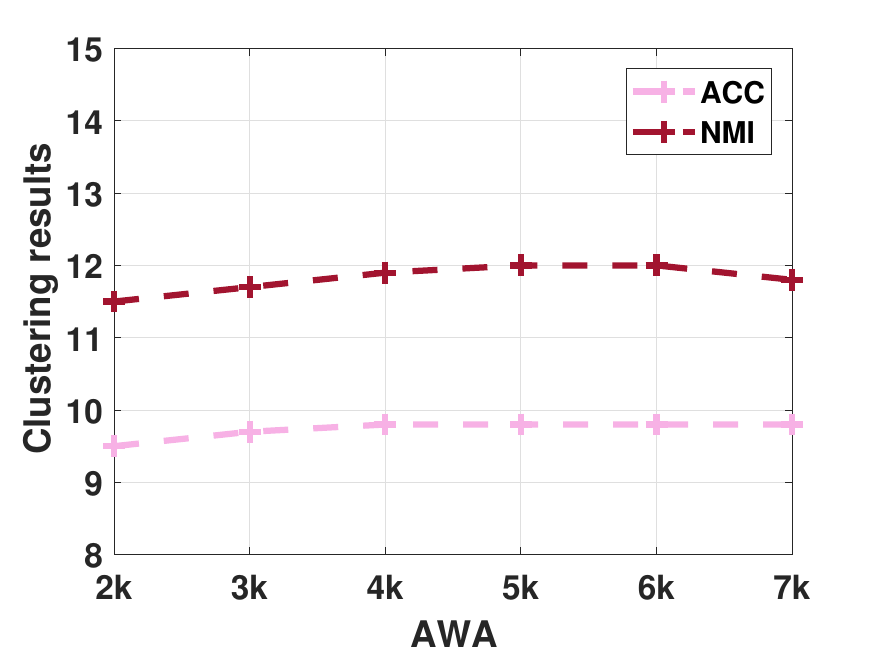}  
		\end{minipage}%
	}
	\subfigure[Caltech-256] 
	{
		\begin{minipage}[t]{0.23\linewidth}
			\centering      
			\includegraphics[width=1.8in]{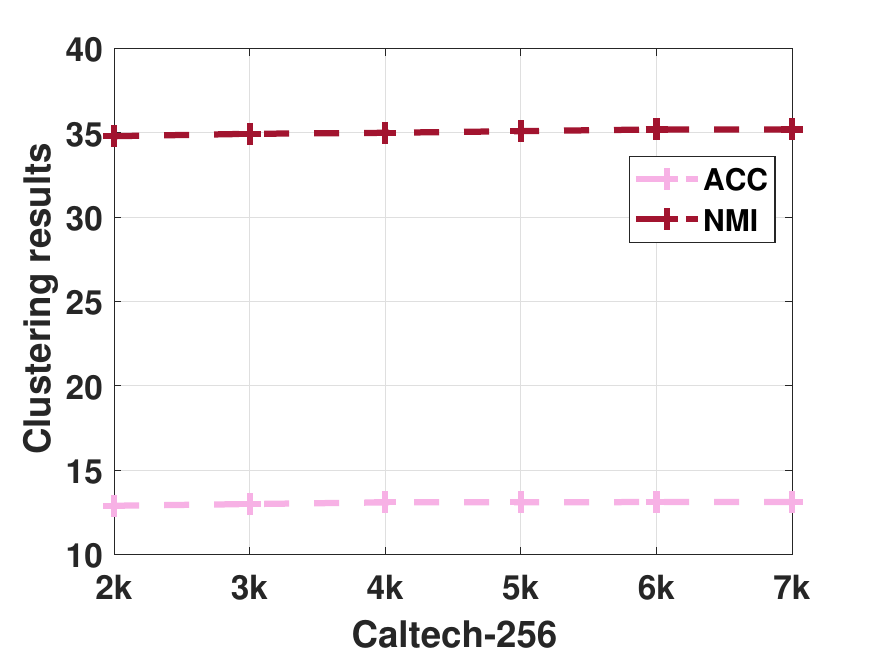} 
		\end{minipage}
	}	
	\subfigure[Flower17] 
	{
		\begin{minipage}[t]{0.23\linewidth}
			\centering      
			\includegraphics[width=1.8in]{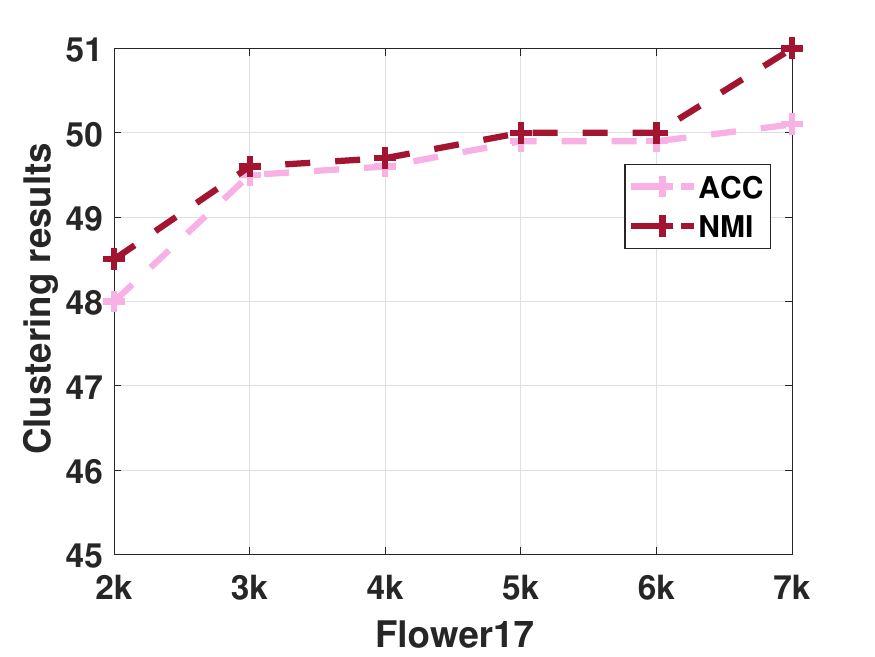} 
		\end{minipage}
	}	
	\subfigure[MNIST] 
	{
		\begin{minipage}[t]{0.23\linewidth}
			\centering          
			\includegraphics[width=1.8in]{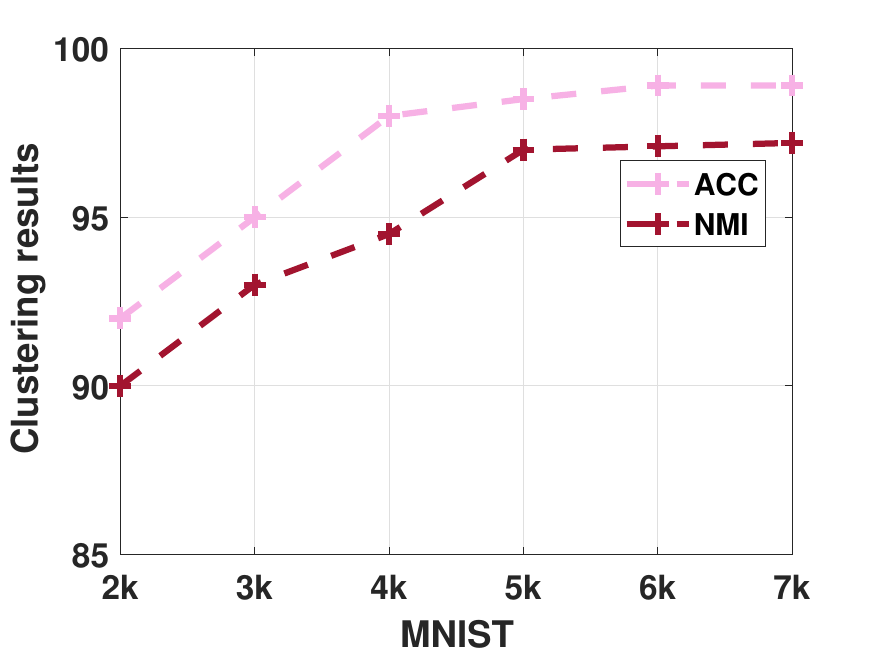}  
		\end{minipage}%
	}
	
	\subfigure[TinyImageNet] 
	{
		\begin{minipage}[t]{0.23\linewidth}
			\centering          
			\includegraphics[width=1.8in]{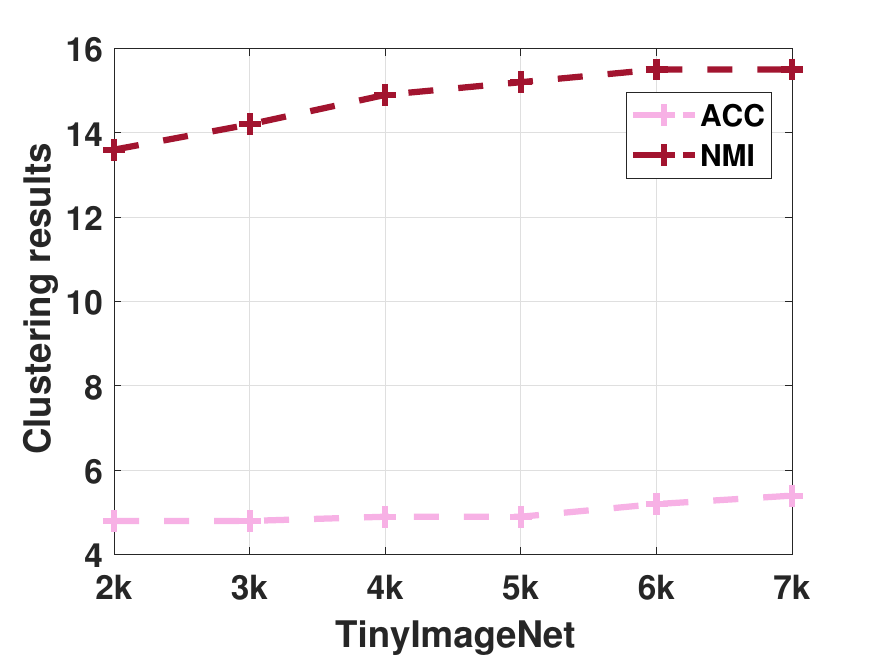}  
		\end{minipage}%
	}
	\subfigure[VGGFace2] 
	{
		\begin{minipage}[t]{0.23\linewidth}
			\centering      
			\includegraphics[width=1.8in]{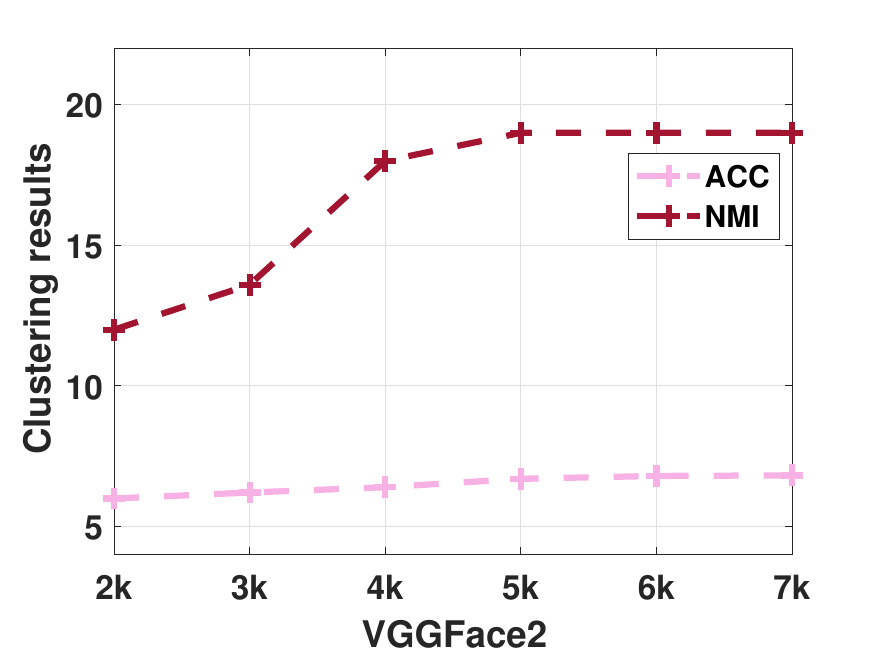} 
		\end{minipage}
	}
	\subfigure[YoutubeFace-50] 
	{
		\begin{minipage}[t]{0.23\linewidth}
			\centering      
			\includegraphics[width=1.8in]{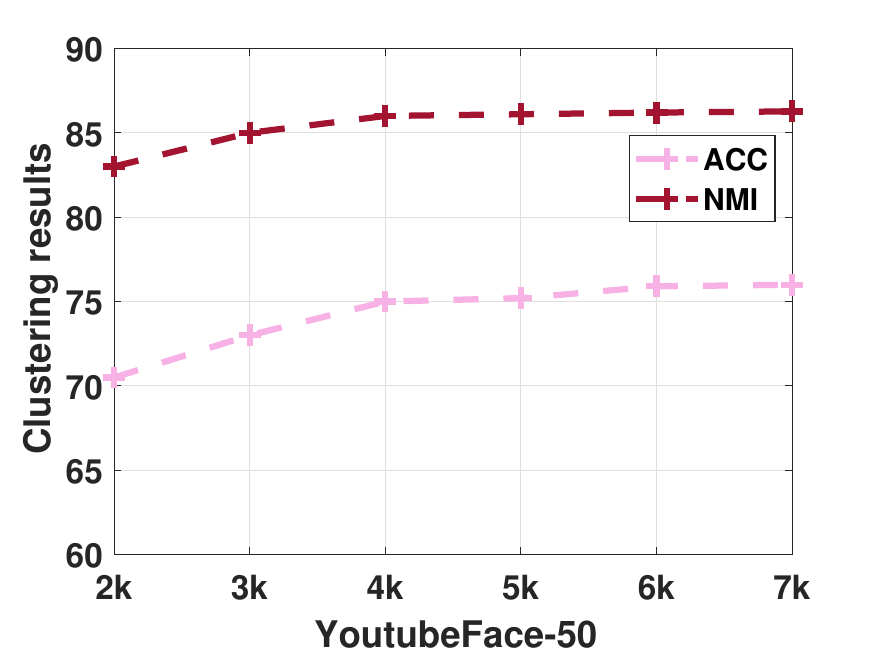} 
		\end{minipage}
	}	
	\caption{Quantitative study of anchors on all datasets in terms of ACC and NMI.} 
	\label{fig1}  
\end{figure*}

\begin{figure*} [!htbp]
	\centering    
	
	\subfigure[AWA] 
	{
		\begin{minipage}[t]{0.23\linewidth}
			\centering          
			\includegraphics[width=1.8in]{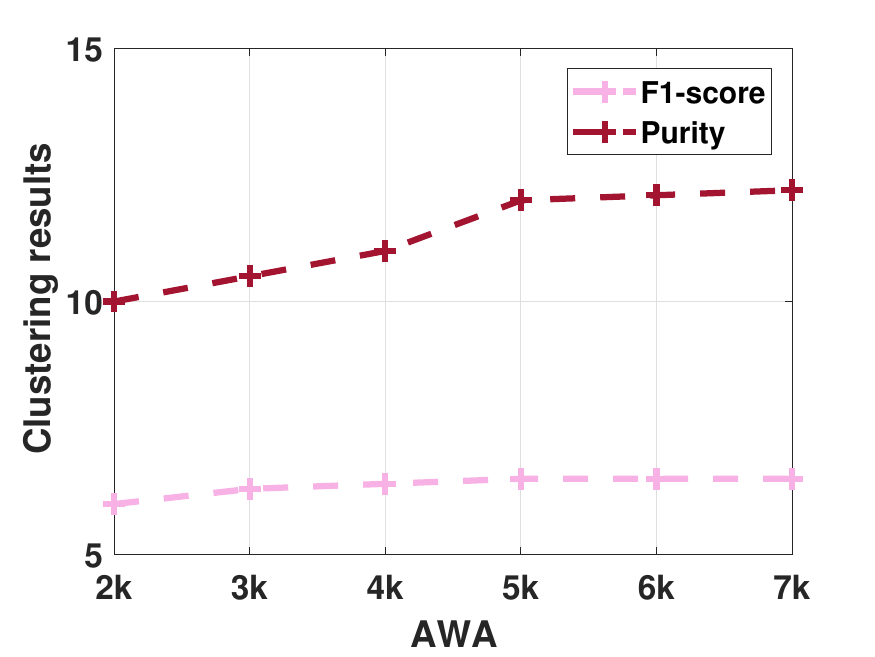}  
		\end{minipage}%
	}
	\subfigure[Caltech-256] 
	{
		\begin{minipage}[t]{0.23\linewidth}
			\centering      
			\includegraphics[width=1.8in]{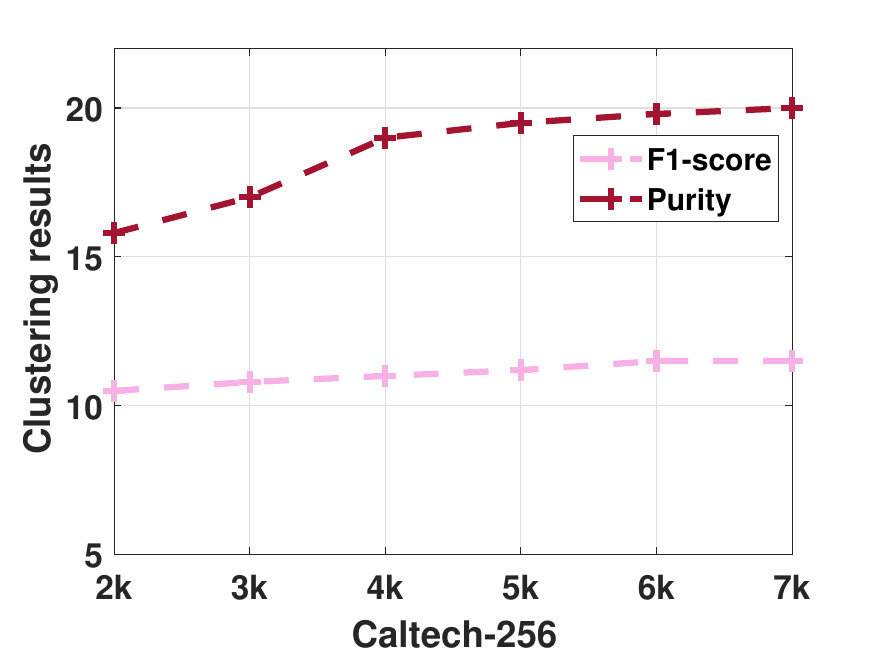} 
		\end{minipage}
	}	
	\subfigure[Flower17] 
	{
		\begin{minipage}[t]{0.23\linewidth}
			\centering      
			\includegraphics[width=1.8in]{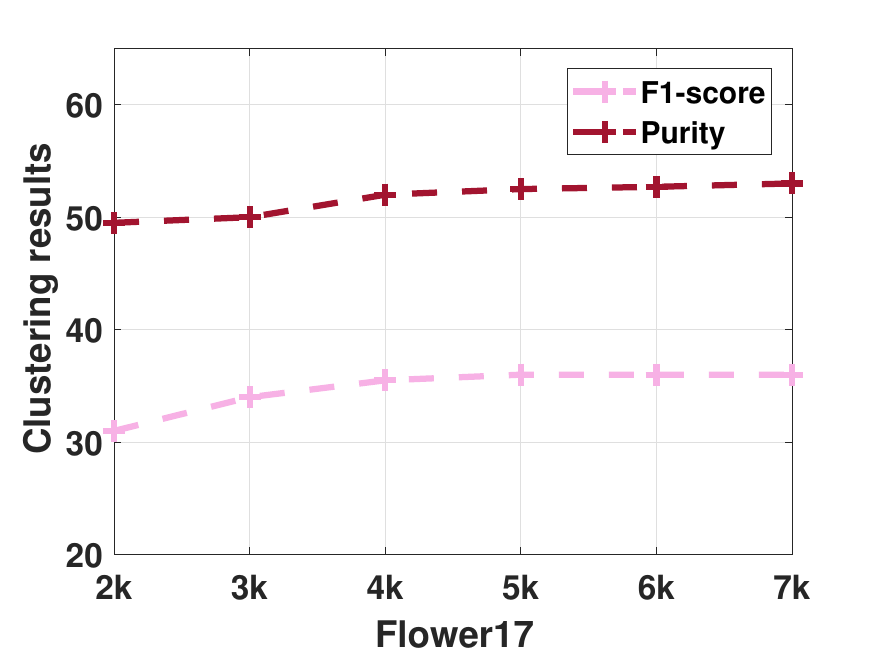} 
		\end{minipage}
	}	
	\subfigure[MNIST] 
	{
		\begin{minipage}[t]{0.23\linewidth}
			\centering          
			\includegraphics[width=1.8in]{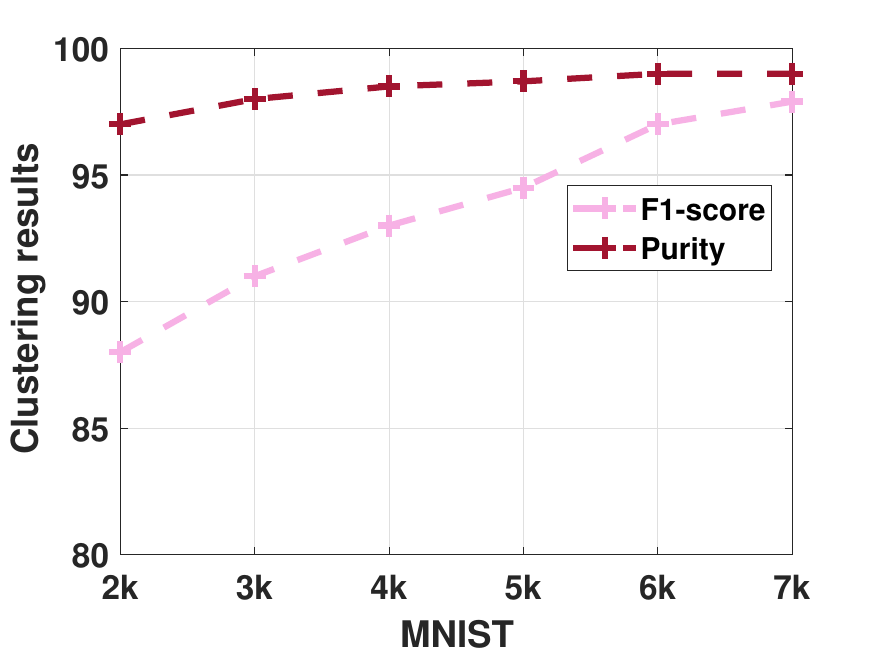}  
		\end{minipage}
	}
	
	\subfigure[TinyImageNet] 
	{
		\begin{minipage}[t]{0.23\linewidth}
			\centering          
			\includegraphics[width=1.8in]{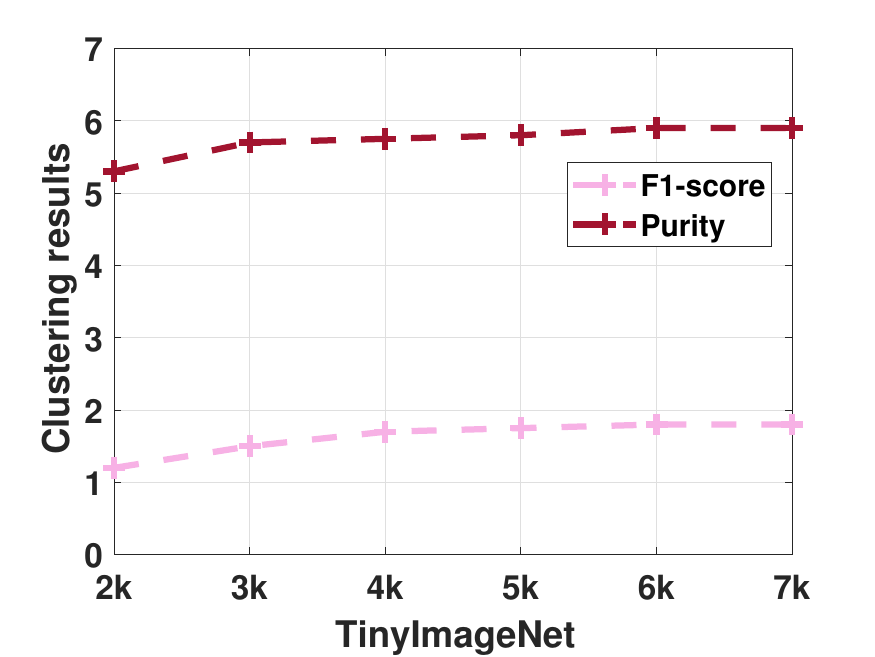}  
		\end{minipage}%
	}
	\subfigure[VGGFace2] 
	{
		\begin{minipage}[t]{0.23\linewidth}
			\centering      
			\includegraphics[width=1.8in]{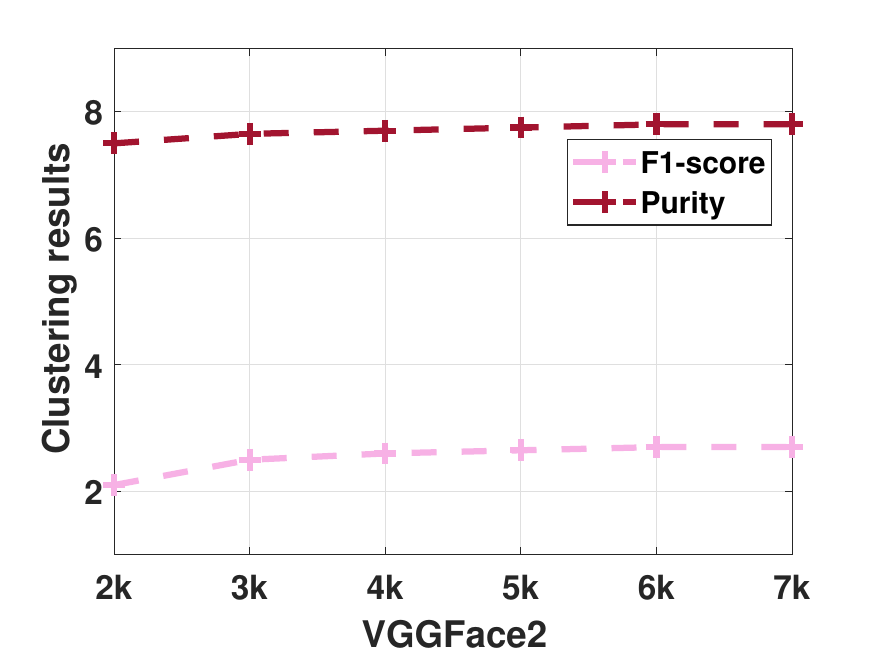} 
		\end{minipage}
	}	
	\subfigure[YoutubeFace-50] 
	{
		\begin{minipage}[t]{0.23\linewidth}
			\centering      
			\includegraphics[width=1.8in]{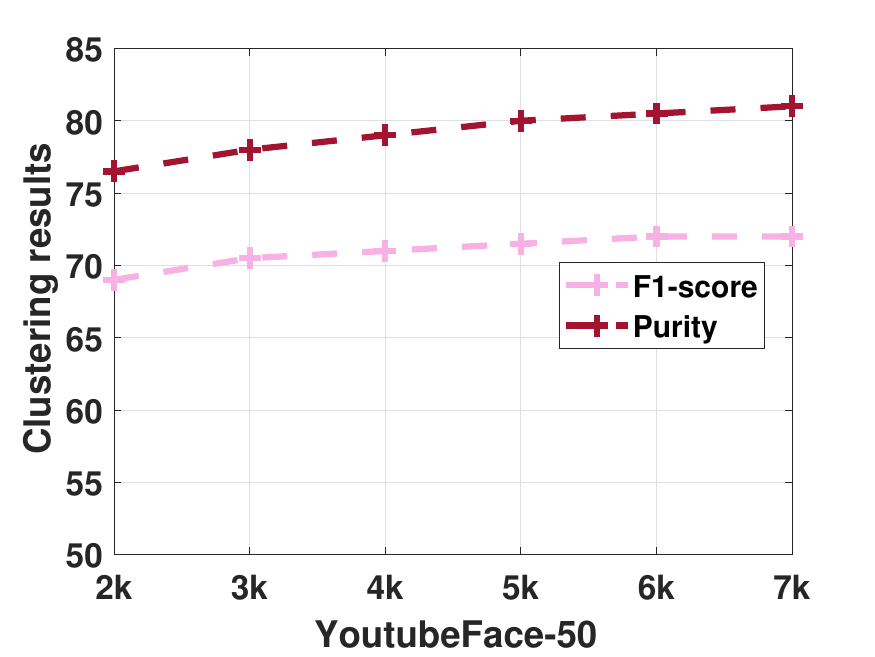} 
		\end{minipage}
	}
	\caption{Quantitative study of anchors on all datasets in terms of F1-score and Purity.} 
	\label{fig1}  
\end{figure*}

\begin{table*}[h] \tiny
	\caption{Running time (s) comparison with representive methods on different datasets. ``-
		" indicates out of memory.}
	\centering
	\tabcolsep=0.56cm
	\begin{tabular}{|c|cccccccc|}
		\toprule[0.5pt]
		\textbf{Dataset}&\textbf{AMGL}&\textbf{BMVC}&\textbf{SMVSC}&\textbf{OPMC}&\textbf{UOMVSC}&\textbf{FPMVSCAG}&\textbf{AWMVC}&\textbf{Ours}\\
		\toprule[0.5pt]
		\textbf{AWA}&-&\textbf{67.80}&1060.00&695.00&-&2834.00&435.00&442.00\\
		\textbf{Caltech-256}&-&\textbf{53.00}&1717.00&212.90&-&2588.00&198.00&800.50\\
		\textbf{Flower17}&380.00&\textbf{1.70}&77.40&45.40&193.00&100.50&43.10&43.50\\
		\textbf{MNIST}&-&\textbf{13.80}&605.00&29.00&-&607.00&563.10&565.00\\
		\textbf{TinyImageNet}&-&\textbf{138.80}&7990.00&972.70&-&5872.00&3160.00&3200.00\\
		\textbf{VGGFace2}&-&\textbf{74.50}&2380.00&499.20&-&4059.20&1735.20&1739.00\\
		\textbf{YoutubeFace-50}&-&\textbf{98.70}&1820.00&101.90&-&2690.20&4162.00&4210.00\\
		\toprule[0.5pt]
	\end{tabular}
	\label{}
\end{table*}
\begin{figure*} [!htbp]
	\centering    
	
	\subfigure[AWA] 
	{
		\begin{minipage}[t]{0.23\linewidth}
			\centering          
			\includegraphics[width=1.8in]{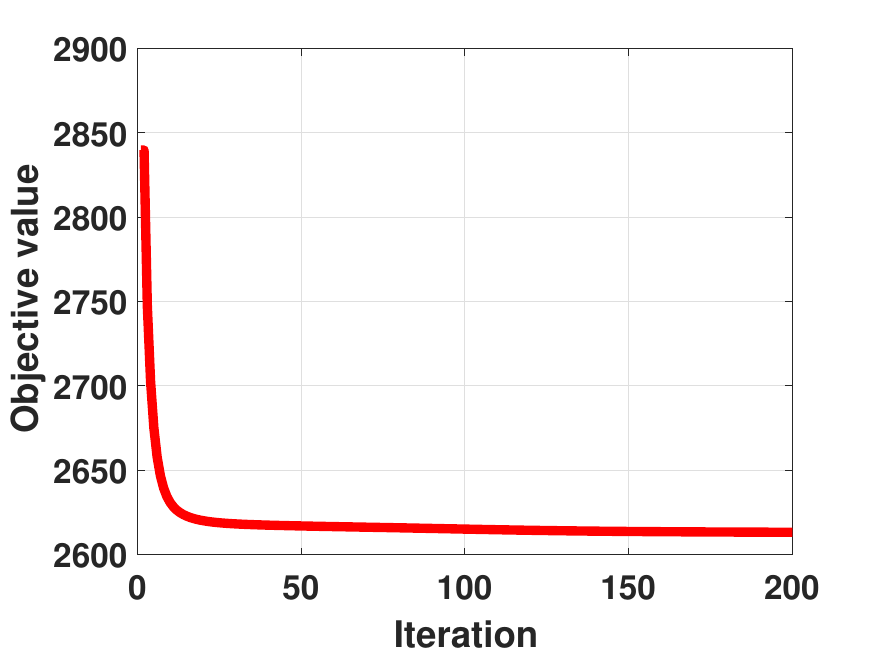}  
		\end{minipage}%
	}
	\subfigure[Caltech-256] 
	{
		\begin{minipage}[t]{0.23\linewidth}
			\centering      
			\includegraphics[width=1.8in]{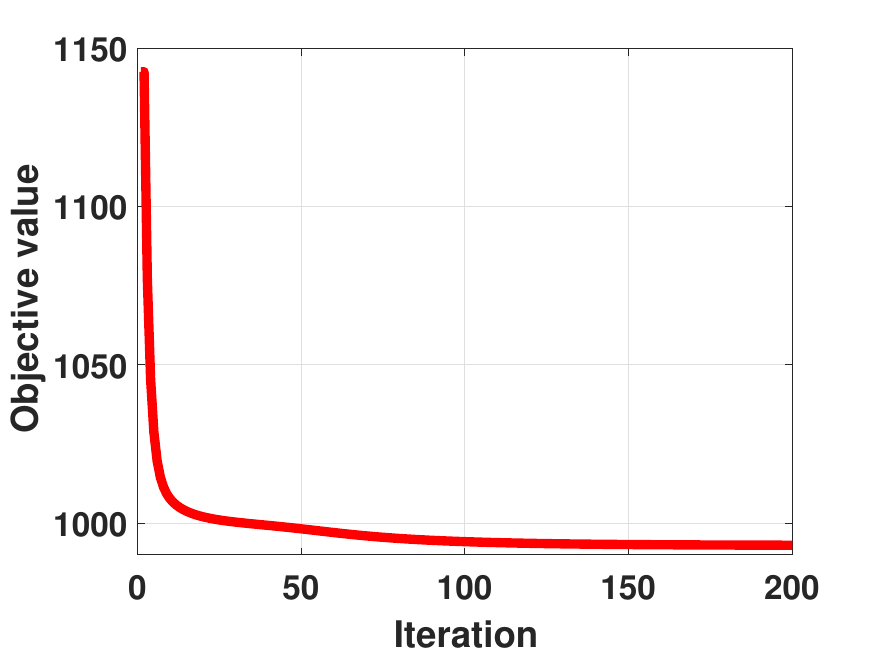} 
		\end{minipage}
	}	
	\subfigure[Flower17] 
	{
		\begin{minipage}[t]{0.23\linewidth}
			\centering      
			\includegraphics[width=1.8in]{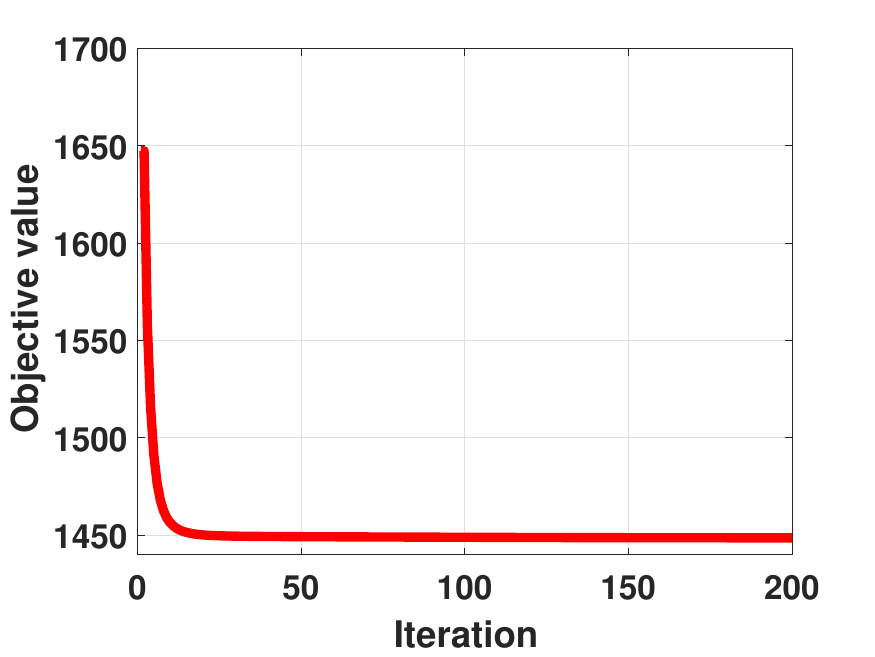} 
		\end{minipage}
	}	
	\subfigure[MNIST] 
	{
		\begin{minipage}[t]{0.23\linewidth}
			\centering          
			\includegraphics[width=1.8in]{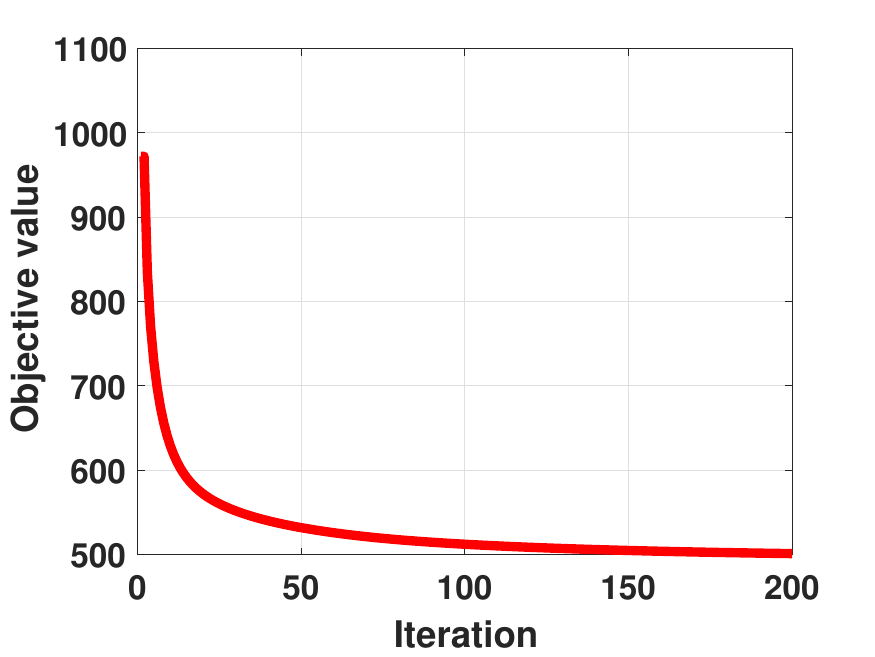}  
		\end{minipage}%
	}
	\caption{The objective values of the proposed method on different datasets with iterations.} 
	\label{fig1}  
\end{figure*}
\section{Experiment}
In this part, we verify the performance of the proposed method by performing comprehensive experiments on different benchmark datasets with large scales, which involves the parameter selection, clustering performance comparison, ablation study, running time comparison, quantitative study of anchors and convergence analysis.

\subsection{Datasets and Compared methods}
We employ seven benchmark datasets for demonstrating the performance of the proposed method, which include AwA \cite{Xinh23}, Caltech-256, Flower17, MNIST, TinyImageNet, VGGFace2 \cite{Xinh23} and YouTubeFace-50. The details of these datasets are shown in Table \uppercase\expandafter{\romannumeral1}.

Our method is compared with seven representive works and these algorithms are summarized as follows: \textbf{AMGL} \cite{Feip16} learns the optimal weights for different graphs and then achieves the global optimal results. \textbf{BMVC} \cite{Zheng19} simultaneously performs discrete representation learning and binary clustering structure learning. \textbf{SMVSC} \cite{Meng21} unifies subspace learning and anchor learning into a framework, which is able to obtain more representative information with the consensus anchors. \textbf{OPMC} \cite{Jiy21} obtains the clustering partition without considering the non-negative constraint. \textbf{UOMVSC} \cite{Chang23} integrates $ K $-means and spectral embedding into a unified framework.
\textbf{FPMVSCAG} \cite{Siw22} simultaneously performs subspace graph construction and anchor selection, which is also parameter-free. \textbf{OMSC} \cite{Man22} achieves a more flexible and representative cluster indicator and anchor representation in a unified framework. \textbf{AWMVC} \cite{Xinh23} learns coefficients with multiple dimensions and then fuses these coefficient matrices, resulting in an optimal consensus matrix. It is able to achieve more comprehensive knowledge and better performance.

In the experiment, we run $ K $-means 20 times to achieve the final clustering results, which eliminates the randomness during the optimization process. For the hyperparameters in the proposed method, we tune them by the grid search strategy and list the best results. We perform experiments on a standard Window PC with AMD Ryzen 5 1600X 3.60 GHz. 

\subsection{Parameter Selection}
In this section, parameter selection regarding $ \beta $ is conducted on all datasets in terms of different metrics. We tune this parameter in $ [0.0001, 0.001, 0.01, 0.1, 1, 10] $ and report the impacts leaded by $ \beta $ on different datasets. According to Figs. 2-3, we observe that superior performance is achieved with $ \beta=0.1 $ in terms of four metrics, which fully shows the validity of considering this term with proper weight in optimization. Moreover, the performance of our method is generally stable for different values on all datasets, showing that our method owns the robustness to $ \beta $.

\subsection{Experimental Results}
The proposed method is compared with seven representive works on different benchmark multi-view datasets under four metrics, which include accuracy (ACC), normalized mutual information (NMI), F1-score and Purity. We give the clustering results and use '-' to indicate that the out-of-memory error is caused for the algorithm. According to Tables. \uppercase\expandafter{\romannumeral2}-\uppercase\expandafter{\romannumeral5}, we observe that

\begin{enumerate}
	\item The proposed method shows excellent clustering performance on different datasets under four metrics. For instance, the proposed method outperforms UOMVSC on Flower17 by 16.09\% in terms of NMI. The superior performance shows the superiority of our method for multi-view clustering.
	
	\item  Compared with other works, anchor-based methods have obvious advantages in clustering performance on multi-view datasets. They learn consensus anchor graph by fusing view-specific information across views and mine more informative feature representation.
	
	\item Among the works based on anchor, our method is still superior in the final clustering performance. FPMVSCAG is a representative of the anchor-based methods, which is comprehensively surpassed by our method on most datasets in performance. It can be explained by the fact that the discriminative anchor learning is able to increase the quality of the shared anchor graph for multi-view clustering and achieve complementarity among different views. Compared with SMVSC, the proposed method considers learning the discriminative feature representation of each view and builds anchors based on these representations. The reason why the proposed method achieves better performance than AWMVC is that we effectively take the discriminative view-specific feature representation learning into consideration, resulting in more desired performance.
	
\end{enumerate}

\subsection{Ablation Study}
In order to demonstrate the validity of adopting the anchor $ A^{p} $ instead of $ H^{p}C $ in Eq. (7), we design the first ablation by replacing $ H^{p}C  $ with $ A^{p} $ in Eq. (9), denoted by ablation-1. Moreover, we conduct the second ablation to validate the effectiveness of imposing orthogonal constraints on $ S $ by applying the constraint $ 0\leq S \leq 1 $ to the $ S $ instead of the constraint $ S^{T}S=I $, denoted by ablation-2.

We report the performance for the proposed method based on ablation experiments including ablation-1 and ablation-2 under four different metrics on Tables \uppercase\expandafter{\romannumeral6}-\uppercase\expandafter{\romannumeral9}. Compared with the clustering results in two ablation experiments, the proposed method has obvious advantages in the final performance, which fully validates the necessarity of using $ A^{p} $ to replace $ H^{p}C $ with the guidance of applying the orthogonal constraints to $ S $ in the optimization process.

\subsection{Running Time Comparison}
The running time for different multi-view clustering approaches on all datasets are listed in this section. According to Table \uppercase\expandafter{\romannumeral10}, it is observed that relatively acceptable running time is consumed by the proposed method, demonstrating that the desired efficiency can be guaranteed. It also further shows that directly learning the anchor $ A^{p} $ for each view instead of adopting $ H^{p}C $ is necessary, which removes the quadratic term regarding $ H^{p} $ in the optimization process and reduces the computation complexity to linear.

\subsection{Quantitative Study of Anchors}
In this part, we investigate how the performance of our method is influenced by the total number of anchors and conduct a quantitative study regarding the number of anchors on all datasets in terms of four metrics. According to Figs. 4-5, it is observed that our method generates the relatively stable performance with varying anchor numbers on datasets.

\subsection{Convergence Analysis}
We have theoretically given the convergence analysis of the proposed method in Section \uppercase\expandafter{\romannumeral3}. To further illustrate the convergence of the algorithm, we perform experiments to show how the objective values change with the number of iterations. From Fig. 6, we find that the target value monotonically decreases as the number of iteration increases, which needs less than 20 iterations to converge.

\section{Conclusion}
This paper proposes discriminative anchor learning for multi-view clustering to increase the quality of the shared anchor graph and achieve complementarity among different views. It formulates discriminative view-specific feature learning and consensus anchor graph construction into a unified framework, where these two parts improve each other to reach the refinement. We learn the optimal anchors of each view and the consensus anchor graph with the orthogonal constraints. Experiments on seven datasets validate the superiority of our method in terms of effectiveness and efficiency. 

\textbf{Acknowledgement} This work was funded by the European Union under NextGeneration EU. Views and opinions expressed are however those of the author(s) only and do not necessarily reflect those of the European Union or The European Research Executive Agency. Neither the European Union nor the granting authority can be held responsible for them.
\bibliographystyle{IEEEtran}

\bibliography{IEEEabrv,IEEEexample}

\begin{thebibliography}{10}
\providecommand{\url}[1]{#1}
\csname url@samestyle\endcsname
\providecommand{\newblock}{\relax}
\providecommand{\bibinfo}[2]{#2}
\providecommand{\BIBentrySTDinterwordspacing}{\spaceskip=0pt\relax}
\providecommand{\BIBentryALTinterwordstretchfactor}{4}
\providecommand{\BIBentryALTinterwordspacing}{\spaceskip=\fontdimen2\font plus
\BIBentryALTinterwordstretchfactor\fontdimen3\font minus
  \fontdimen4\font\relax}
\providecommand{\BIBforeignlanguage}[2]{{%
\expandafter\ifx\csname l@#1\endcsname\relax
\typeout{** WARNING: IEEEtran.bst: No hyphenation pattern has been}%
\typeout{** loaded for the language `#1'. Using the pattern for}%
\typeout{** the default language instead.}%
\else
\language=\csname l@#1\endcsname
\fi
#2}}
\providecommand{\BIBdecl}{\relax}
\BIBdecl

\bibitem{Yuans24}
Y.~Sun, Z.~Ren, P.~Hu, D.~Peng, and X.~Wang, ``Hierarchical consensus hashing
  for cross-modal retrieval,'' \emph{{IEEE} Trans. Multim.}, vol.~26, pp.
  824--836, 2024.

\bibitem{Yuans23}
Y.~Sun, X.~Wang, D.~Peng, Z.~Ren, and X.~Shen, ``Hierarchical hashing learning
  for image set classification,'' \emph{{IEEE} Trans. Image Process.}, vol.~32,
  pp. 1732--1744, 2023.

\bibitem{Yangq23}
Y.~Qin, Y.~Sun, D.~Peng, J.~T. Zhou, X.~Peng, and P.~Hu, ``Cross-modal active
  complementary learning with self-refining correspondence,'' in \emph{Advances
  in Neural Information Processing Systems}, 2023.

\bibitem{Yangq231}
Y.~Qin, Y.~Chen, D.~Peng, X.~Peng, J.~T. Zhou, and P.~Hu,
  ``Noisy-correspondence learning for text-to-image person re-identification,''
  \emph{CoRR}, vol. abs/2308.09911, 2023.

\bibitem{Yala222}
Y.~Qin, H.~Wu, J.~Zhao, and G.~Feng, ``Enforced block diagonal subspace
  clustering with closed form solution,'' \emph{Pattern Recognit.}, vol. 130,
  p. 108791, 2022.

\bibitem{Yala211}
Y.~Qin, H.~Wu, and G.~Feng, ``Structured subspace learning-induced symmetric
  nonnegative matrix factorization,'' \emph{Signal Process.}, vol. 186, p.
  108115, 2021.

\bibitem{Yala232}
Y.~Qin, X.~Zhang, L.~Shen, and G.~Feng, ``Maximum block energy guided robust
  subspace clustering,'' \emph{{IEEE} Trans. Pattern Anal. Mach. Intell.},
  vol.~45, no.~2, pp. 2652--2659, 2023.

\bibitem{Yala233}
Y.~Qin, G.~Feng, Y.~Ren, and X.~Zhang, ``Block-diagonal guided symmetric
  nonnegative matrix factorization,'' \emph{{IEEE} Trans. Knowl. Data Eng.},
  vol.~35, no.~3, pp. 2313--2325, 2023.

\bibitem{Nanp23}
N.~Pu, Z.~Zhong, X.~Ji, and N.~Sebe, ``Federated generalized category
  discovery,'' \emph{CoRR}, vol. abs/2305.14107, 2023.

\bibitem{Peng23}
Z.~Peng, H.~Liu, Y.~Jia, and J.~Hou, ``Deep attention-guided graph clustering
  with dual self-supervision,'' \emph{IEEE Transactions on Circuits and Systems
  for Video Technology}, vol.~33, no.~7, pp. 3296--3307, 2023.

\bibitem{Peng22}
Z.~Peng, Y.~Jia, H.~Liu, J.~Hou, and Q.~Zhang, ``Maximum entropy subspace
  clustering network,'' \emph{IEEE Transactions on Circuits and Systems for
  Video Technology}, vol.~32, no.~4, pp. 2199--2210, 2022.

\bibitem{Wuw21}
W.~Wu, Y.~Jia, S.~Wang, R.~Wang, H.~Fan, and S.~Kwong, ``Positive and negative
  label-driven nonnegative matrix factorization,'' \emph{IEEE Transactions on
  Circuits and Systems for Video Technology}, vol.~31, no.~7, pp. 2698--2710,
  2021.

\bibitem{Jia23}
Y.~Jia, G.~Lu, H.~Liu, and J.~Hou, ``Semi-supervised subspace clustering via
  tensor low-rank representation,'' \emph{IEEE Transactions on Circuits and
  Systems for Video Technology}, vol.~33, no.~7, pp. 3455--3461, 2023.

\bibitem{Yala24}
Y.~Qin, N.~Pu, and H.~Wu, ``Edmc: Efficient multi-view clustering via cluster
  and instance space learning,'' \emph{IEEE Transactions on Multimedia},
  vol.~26, pp. 5273--5283, 2024.

\bibitem{Chenyo22}
Y.~Chen, S.~Wang, X.~Xiao, Y.~Liu, Z.~Hua, and Y.~Zhou, ``Self-paced enhanced
  low-rank tensor kernelized multi-view subspace clustering,'' \emph{IEEE
  Transactions on Multimedia}, vol.~24, pp. 4054--4066, 2022.

\bibitem{Yala221}
Y.~Qin, H.~Wu, X.~Zhang, and G.~Feng, ``Semi-supervised structured subspace
  learning for multi-view clustering,'' \emph{{IEEE} Trans. Image Process.},
  vol.~31, pp. 1--14, 2022.

\bibitem{Yala231}
Y.~Qin, G.~Feng, Y.~Ren, and X.~Zhang, ``Consistency-induced multiview subspace
  clustering,'' \emph{{IEEE} Trans. Cybern.}, vol.~53, no.~2, pp. 832--844,
  2023.

\bibitem{Yala241}
Y.~Qin, N.~Pu, and H.~Wu, ``Elastic multi-view subspace clustering with
  pairwise and high-order correlations,'' \emph{{IEEE} Trans. Knowl. Data
  Eng.}, vol.~36, no.~2, pp. 556--568, 2024.

\bibitem{Yala234}
Y.~Qin, C.~Qin, X.~Zhang, D.~Qi, and G.~Feng, ``Nim-nets: Noise-aware
  incomplete multi-view learning networks,'' \emph{{IEEE} Trans. Image
  Process.}, vol.~32, pp. 175--189, 2023.

\bibitem{Yala242}
Y.~Qin, Z.~Tang, H.~Wu, and G.~Feng, ``Flexible tensor learning for multi-view
  clustering with markov chain,'' \emph{{IEEE} Trans. Knowl. Data Eng.},
  vol.~36, no.~4, pp. 1552--1565, 2024.

\bibitem{Xingf23}
X.~Li, Y.~Sun, Q.~Sun, Z.~Ren, and Y.~Sun, ``Cross-view graph matching guided
  anchor alignment for incomplete multi-view clustering,'' \emph{Inf. Fusion},
  vol. 100, p. 101941, 2023.

\bibitem{Qian18}
Q.~Wang, Z.~Ding, Z.~Tao, Q.~Gao, and Y.~Fu, ``Partial multi-view clustering
  via consistent {GAN},'' in \emph{{IEEE} International Conference on Data
  Mining}, 2018, pp. 1290--1295.

\bibitem{Qian21}
Q.~Wang, J.~Cheng, Q.~Gao, G.~Zhao, and L.~Jiao, ``Deep multi-view subspace
  clustering with unified and discriminative learning,'' \emph{{IEEE} Trans.
  Multim.}, vol.~23, pp. 3483--3493, 2021.

\bibitem{Qian211}
Q.~Wang, Z.~Ding, Z.~Tao, Q.~Gao, and Y.~Fu, ``Generative partial multi-view
  clustering with adaptive fusion and cycle consistency,'' \emph{{IEEE} Trans.
  Image Process.}, vol.~30, pp. 1771--1783, 2021.

\bibitem{Tie23}
T.~Zhang, X.~Liu, L.~Gong, S.~Wang, X.~Niu, and L.~Shen, ``Late fusion multiple
  kernel clustering with local kernel alignment maximization,'' \emph{{IEEE}
  Trans. Multim.}, vol.~25, pp. 993--1007, 2023.

\bibitem{Ghu19}
G.~A. Khan, J.~Hu, T.~Li, B.~Diallo, and Q.~Huang, ``Weighted multi-view data
  clustering via joint non-negative matrix factorization,'' in \emph{{IEEE}
  International Conference on Intelligent Systems and Knowledge Engineering},
  2019, pp. 1159--1165.

\bibitem{Hong15}
H.~Gao, F.~Nie, X.~Li, and H.~Huang, ``Multi-view subspace clustering,'' in
  \emph{{IEEE} International Conference on Computer Vision}, 2015, pp.
  4238--4246.

\bibitem{Weix22}
W.~Liang, S.~Zhou, J.~Xiong, X.~Liu, S.~Wang, E.~Zhu, Z.~Cai, and X.~Xu,
  ``Multi-view spectral clustering with high-order optimal neighborhood
  laplacian matrix,'' \emph{{IEEE} Trans. Knowl. Data Eng.}, vol.~34, no.~7,
  pp. 3418--3430, 2022.

\bibitem{Liux21}
X.~Liu, ``Incomplete multiple kernel alignment maximization for clustering,''
  \emph{IEEE Transactions on Pattern Analysis and Machine Intelligence}, pp.
  1--1, 2021.

\bibitem{Jing13}
J.~Gao, J.~Han, J.~Liu, and C.~Wang, ``Multi-view clustering via joint
  nonnegative matrix factorization,'' in \emph{Proceedings of the {SIAM}
  International Conference on Data Mining}, 2013, pp. 252--260.

\bibitem{Xuel22}
X.~Li, H.~Zhang, R.~Wang, and F.~Nie, ``Multiview clustering: {A} scalable and
  parameter-free bipartite graph fusion method,'' \emph{{IEEE} Trans. Pattern
  Anal. Mach. Intell.}, vol.~44, no.~1, pp. 330--344, 2022.

\bibitem{Zhao20}
Z.~Kang, W.~Zhou, Z.~Zhao, J.~Shao, M.~Han, and Z.~Xu, ``Large-scale multi-view
  subspace clustering in linear time,'' in \emph{The {AAAI} Conference on
  Artificial Intelligence}, 2020, pp. 4412--4419.

\bibitem{Yeq15}
Y.~Li, F.~Nie, H.~Huang, and J.~Huang, ``Large-scale multi-view spectral
  clustering via bipartite graph,'' in \emph{Proceedings of the {AAAI}
  Conference on Artificial Intelligence}, 2015, pp. 2750--2756.

\bibitem{Suy22}
S.~Liu, S.~Wang, P.~Zhang, K.~Xu, X.~Liu, C.~Zhang, and F.~Gao, ``Efficient
  one-pass multi-view subspace clustering with consensus anchors,'' in
  \emph{Proceedings of the {AAAI} Conference on Artificial Intelligence}, 2022,
  pp. 7576--7584.

\bibitem{Jiy21}
J.~Liu, X.~Liu, Y.~Yang, L.~Liu, S.~Wang, W.~Liang, and J.~Shi, ``One-pass
  multi-view clustering for large-scale data,'' in \emph{{IEEE/CVF}
  International Conference on Computer Vision}, 2021, pp. 12\,324--12\,333.

\bibitem{Jing18}
J.~Wang, F.~Tian, H.~Yu, C.~H. Liu, K.~Zhan, and X.~Wang, ``Diverse
  non-negative matrix factorization for multiview data representation,''
  \emph{{IEEE} Trans. Cybern.}, vol.~48, no.~9, pp. 2620--2632, 2018.

\bibitem{Siw22}
S.~Wang, X.~Liu, X.~Zhu, P.~Zhang, Y.~Zhang, F.~Gao, and E.~Zhu, ``Fast
  parameter-free multi-view subspace clustering with consensus anchor
  guidance,'' \emph{{IEEE} Trans. Image Process.}, vol.~31, pp. 556--568, 2022.

\bibitem{Xue22}
X.~Li, H.~Zhang, R.~Wang, and F.~Nie, ``Multiview clustering: {A} scalable and
  parameter-free bipartite graph fusion method,'' \emph{{IEEE} Trans. Pattern
  Anal. Mach. Intell.}, vol.~44, no.~1, pp. 330--344, 2022.

\bibitem{Wangj23}
J.~Wang, C.~Tang, X.~Zheng, X.~Liu, W.~Zhang, E.~Zhu, and X.~Zhu, ``Fast
  approximated multiple kernel k-means,'' \emph{IEEE Transactions on Knowledge
  and Data Engineering}, pp. 1--10, 2023.

\bibitem{Chenyo23}
Y.~Chen, X.~Zhao, Z.~Zhang, Y.~Liu, J.~Su, and Y.~Zhou, ``Tensor learning meets
  dynamic anchor learning: From complete to incomplete multiview clustering,''
  \emph{IEEE Transactions on Neural Networks and Learning Systems}, pp. 1--14,
  2023.

\bibitem{Wangju23}
J.~Wang, C.~Tang, Z.~Wan, W.~Zhang, K.~Sun, and A.~Y. Zomaya, ``Efficient and
  effective one-step multiview clustering,'' \emph{IEEE Transactions on Neural
  Networks and Learning Systems}, pp. 1--12, 2023.

\bibitem{Huangbo20}
B.~Huang, T.~Xu, S.~Jiang, Y.~Chen, and Y.~Bai, ``Robust visual tracking via
  constrained multi-kernel correlation filters,'' \emph{IEEE Transactions on
  Multimedia}, vol.~22, no.~11, pp. 2820--2832, 2020.

\bibitem{Meng21}
M.~Sun, P.~Zhang, S.~Wang, S.~Zhou, W.~Tu, X.~Liu, E.~Zhu, and C.~Wang,
  ``Scalable multi-view subspace clustering with unified anchors,'' in
  \emph{{ACM} Multimedia Conference}, 2021, pp. 3528--3536.

\bibitem{Fei23}
F.~Nie, W.~Chang, R.~Wang, and X.~Li, ``Learning an optimal bipartite graph for
  subspace clustering via constrained laplacian rank,'' \emph{{IEEE} Trans.
  Cybern.}, vol.~53, no.~2, pp. 1235--1247, 2023.

\bibitem{Man22}
M.~Chen, C.~Wang, D.~Huang, J.~Lai, and P.~S. Yu, ``Efficient orthogonal
  multi-view subspace clustering,'' in \emph{{ACM} {SIGKDD} Conference on
  Knowledge Discovery and Data Mining}, 2022, pp. 127--135.

\bibitem{Tiej22}
T.~Zhang, X.~Liu, E.~Zhu, S.~Zhou, and Z.~Dong, ``Efficient anchor
  learning-based multi-view clustering - {A} late fusion method,'' in
  \emph{{ACM} International Conference on Multimedia}, 2022, pp. 3685--3693.

\bibitem{Feip17}
F.~Nie, J.~Li, and X.~Li, ``Self-weighted multiview clustering with multiple
  graphs,'' in \emph{Proceedings of the International Joint Conference on
  Artificial Intelligence}, 2017, pp. 2564--2570.

\bibitem{Xinh23}
X.~Wan, X.~Liu, J.~Liu, S.~Wang, Y.~Wen, W.~Liang, E.~Zhu, Z.~Liu, and L.~Zhou,
  ``Auto-weighted multi-view clustering for large-scale data,'' \emph{CoRR},
  vol. abs/2303.01983, 2023.

\bibitem{Feip16}
F.~Nie, J.~Li, and X.~Li, ``Parameter-free auto-weighted multiple graph
  learning: {A} framework for multiview clustering and semi-supervised
  classification,'' in \emph{Proceedings of the International Joint Conference
  on Artificial Intelligence}, 2016, pp. 1881--1887.

\bibitem{Zheng19}
Z.~Zhang, L.~Liu, F.~Shen, H.~T. Shen, and L.~Shao, ``Binary multi-view
  clustering,'' \emph{{IEEE} Trans. Pattern Anal. Mach. Intell.}, vol.~41,
  no.~7, pp. 1774--1782, 2019.

\bibitem{Chang23}
C.~Tang, Z.~Li, J.~Wang, X.~Liu, W.~Zhang, and E.~Zhu, ``Unified one-step
  multi-view spectral clustering,'' \emph{{IEEE} Trans. Knowl. Data Eng.},
  vol.~35, no.~6, pp. 6449--6460, 2023.

\end{thebibliography}

\begin{IEEEbiography}
	[{\includegraphics[width=1in,height=1.25in,clip,keepaspectratio]{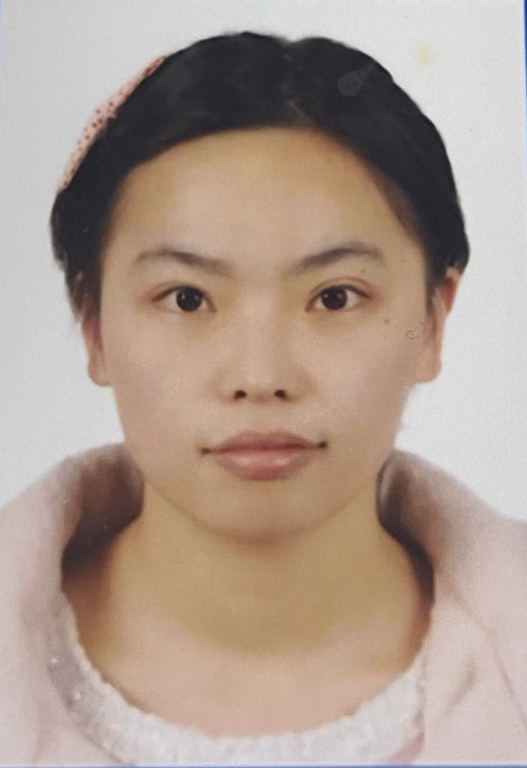}}]{Yalan Qin}
	received the PhD degree from Shanghai University in 2022. Currently, she is a research fellow in Shanghai University, Shanghai, China. Her current research interests include multi-view learning, pattern recognition and image processing. She has published papers in journals and conferences, including IEEE TPAMI, IEEE TIP, IEEE TKDE, IEEE TMM, CVPR, and ACM MM etc.
\end{IEEEbiography}

\begin{IEEEbiography}
	[{\includegraphics[width=1in,height=1.25in,clip,keepaspectratio]{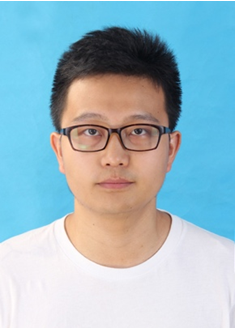}}]{Nan Pu}
	received his Ph.D. degree at Leiden University in the Netherlands in 2022. He is currently working as a postdoctoral researcher at the University of Trento, Italy. His research interests focus on lifelong learning and multi-modality learning with deep learning methods. He has published papers in international journals and conferences, including IEEE TPAMI, IEEE TKDE, IEEE TMM, CVPR, AAAI, ACM MM, and ICASSP etc.
\end{IEEEbiography}

\begin{IEEEbiography}
	[{\includegraphics[width=1in,height=1.25in,clip,keepaspectratio]{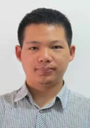}}]{Hanzhou Wu (M'16)}
	received the B.S. and Ph.D. degrees from Southwest Jiaotong University, Chengdu, China, in June 2011 and June 2017, respectively. From October 2014 to October 2016, he was a Visiting Scholar with the New Jersey Institute of Technology, NJ, USA. He was a Research Staff with the Institute of Automation, Chinese Academy of Sciences, Beijing, China, from July 2017 to February 2019. He is currently an Associate Professor with Shanghai University, Shanghai, China. His research interests include steganography, steganalysis and digital watermarking. He has authored/co-authored more than 50 research papers and 4 book chapters. He served as the Local Organization Chair of 14th IEEE International Workshop on Information Forensics and Security, the Steering Committee Member of 14th/15th/16th International Conference on Advances in Multimedia, and the Technical Committee Member of Multimedia Security and Forensics of Asia-Pacific Signal and Information Processing Association (APSIPA). He was awarded 2022 CCF-Tencent Rhino-Bird Young Faculty Open Research Fund.
\end{IEEEbiography}

\begin{IEEEbiography}
	[{\includegraphics[width=1in,height=1.25in,clip,keepaspectratio]{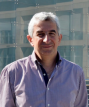}}]{Nicu Sebe}
is professor with the University of Trento, Italy. His research covers the areas of multimedia information retrieval and human behavior understanding. He was the Program Chair of the International Conference on Image and Video Retrieval in 2007 and 2010, ACM Multimedia 2007 and 2011. He was the Program Chair of ICCV 2017 and ECCV 2016, and a General Chair of ACM ICMR 2017.
\end{IEEEbiography}
\end{document}